\documentclass[11pt,a4paper,hyphens,dvipsnames]{article}
\usepackage{authblk}
\usepackage[hyperref]{acl2020}
\usepackage{times}
\usepackage{latexsym}
\usepackage[T1]{fontenc}
\usepackage[utf8]{inputenc}
\usepackage{microtype}
\usepackage{graphicx}
\usepackage{xspace}
\usepackage{xcolor}
\usepackage{txfonts}
\usepackage{scalefnt}
\usepackage{url}
\usepackage{soul}

\usepackage{microtype}

\aclfinalcopy 

\newcommand{\textmc}[1]{\textsc{\scalefont{1.25}#1}}

\newcommand\da{\textmc{da}\xspace}
\newcommand\qa{\textmc{qa}\xspace}
\newcommand\bioasq{\textmc{bioasq}\xspace}
\newcommand\squad{\textmc{squad}\xspace}
\newcommand\squadtwo{\textmc{squad-v}2\xspace}
\newcommand\mlp{\textmc{mlp}\xspace}
\newcommand\albert{\textmc{albert}\xspace}
\newcommand\albertxl{\textmc{albert-xl}\xspace}
\newcommand\distilbert{\textmc{distilbert}\xspace}
\newcommand\bert{\textmc{bert}\xspace}
\newcommand\biobert{\textmc{biobert}\xspace}
\newcommand\bmtf{\textmc{bm}25\xspace}
\newcommand\mrc{\textmc{mrc}\xspace}
\newcommand\biomrc{\textmc{biomrc}\xspace}

\newcommand\softmax{$\mathrm{softmax}$\xspace}
\newcommand\sigmoid{$\mathrm{sigmoid}$\xspace}
\newcommand\nlp{\textmc{nlp}\xspace}
\newcommand\nli{\textmc{nli}\xspace}
\newcommand\pubmed{\textmc{pubmed}\xspace}
\newcommand\ir{\textmc{ir}\xspace}
\newcommand\wtv{\textmc{word}2\textmc{vec}\xspace}

\newcommand\roberta{\textmc{roberta}\xspace}
\newcommand\prauc{\textmc{prauc}\xspace}
\newcommand\auc{\textmc{auc}\xspace}

\newcommand\btr{\textmc{btr}\xspace}
\newcommand\biolm{\textmc{biolm}\xspace}

\newcommand\pmc{\textmc{pmc}\xspace}
\newcommand\mimicthree{\textmc{mimic-iii}\xspace}
\newcommand\qg{\textmc{qg}\xspace}
\newcommand\tfive{\textmc{t}5}
\newcommand\tfivesp{\textmc{t}5\xspace}
\newcommand\context{\textmc{context}\xspace}
\newcommand\fr{\textmc{fr}}
\newcommand\es{\textmc{es}}
\newcommand\de{\textmc{de}}

\title{Data Augmentation for Biomedical Factoid Question Answering}

\author[1,2]{Dimitris Pappas}
\author[1,3]{Prodromos Malakasiotis}
\author[1]{Ion Androutsopoulos}
\affil[1]{Department of Informatics, Athens University of Economics and Business, Greece}
\affil[1]{\url{pappasd@aueb.gr, rulller@aueb.gr, ion@aueb.gr}}
\affil[2]{Institute for Language and Speech Processing, Research Center `Athena', Greece}
\affil[2]{\url{dpappas@athenarc.gr}}
\affil[3]{Institute of Informatics and Telecommunications, NCSR `Demokritos', Greece}

\date{}

\begin{document}

\maketitle

\sethlcolor{Goldenrod}

\begin{abstract}

We study the effect of seven data augmentation (\da) methods in factoid question answering, focusing on the biomedical domain, where obtaining training instances is particularly difficult.
We experiment with data from the \bioasq challenge, which we augment with training instances obtained from an artificial biomedical machine reading comprehension dataset, or via back-translation, information retrieval, word substitution based on \wtv embeddings or masked language modeling, question generation, or extending the given passage with additional context. We show that \da can lead to very significant performance gains, even when using large pre-trained Transformers, contributing to a broader discussion of if/when \da benefits large pre-trained models. 
One of the simplest \da methods, \wtv-based word substitution, performed best and is recommended. 
We release our artificial training instances and code.
\end{abstract}

\section{Introduction} \label{sec:introduction}

Question Answering (\qa) systems aim to answer natural language questions by searching in structured \cite{Fu2020ASO,luo-etal-2018-knowledge,yadati-etal-2021-knowledge} or unstructured data, such as free-text documents \cite{aghaebrahimian-2018-linguistically}. Here we consider \qa of the latter kind. Fully fledged \qa systems for document collections 
retrieve relevant documents, identify relevant passages,  extract and aggregate answer spans 
etc.\ \cite{dr_qa_2017,pappas-androutsopoulos-2021-neural}. There are also different types of questions, e.g., \textit{yes/no}, \textit{factoid}, \textit{list}, \textit{how-to}. Thus, creating realistic datasets to train and evaluate complete \qa systems for document collections is resource intensive, especially for systems targeting specialized domains.
A prime example is the \emph{biomedical domain}, the focus of this work, where obtaining realistic training (and test) instances requires medical expertise, which is costly and difficult to obtain. Consequently, biomedical datasets for full \qa systems contain just a few thousand training instances \cite{Tsatsaronis_et_al_bioasq,moller-2020-COVID-QA} or focus on simpler question types only, e.g., yes/no questions \cite{jin-2019-pubmedqa}.

A simplified form of \qa for textual data is Machine Reading Comprehension (\mrc) \cite{yang-2015-wikiqa,rajpurkar-2016-squad,Campos_2016_MSMARCO,chen_2017_DrQA,lai-2017-race,joshi_2017_triviaqa,kwiatkowski-2019-natural,reddy-2019-coqa,jin-2019-pubmedqa,Wang_2020_ReCO}, where the system is given a question and a particular (or a few) passage(s) and the answer must be found therein. In effect, \mrc focuses on a particular core stage of a full \qa pipeline, identifying 
answer spans, assuming that relevant documents and passages have already been identified. We also focus on this stage, 
adopting an \mrc setting. Large generic (non domain-specific) \mrc datasets have been constructed via crowd-annotation \cite{rajpurkar-2016-squad,rajpurkar-etal-2018-squad-v2,yang-choi-2019-friendsqa,joshi_2017_triviaqa}, but crowd-annotation on that scale is difficult when biomedical expertise is required. An alternative is to \emph{automatically} generate \emph{cloze-style} \mrc datasets. 
The last sentence or title of a random passage is treated as a question, some part (e.g., named entity) of the `question' is masked, and the system is required to predict it.
This approach has been used to generate large \emph{artificial} cloze-style \mrc datasets 
\cite{Hill_et_al_CBTest,chen-etal-2016-CNN-daily,Bajgar_et_al_BookTest}, including biomedical ones 
\citep{pappas-2018-bioread,pappas-2020-biomrc}. These datasets could be used to augment real ones,
but have mostly been used as artificial experimental setups only.  

When  training examples for end-tasks are limited, as in realistic biomedical \qa datasets, the currently dominant approach in \nlp is to use pre-trained Transformers \cite{bert,roberta,albert,deberta,t5}, possibly pre-trained on domain-specific corpora \cite{biobert,scibert,legalbert}, and fine-tune (further train) them on the limited examples of the end-tasks. Nevertheless, increasing the number of end-task  examples typically improves performance. One way to achieve this is to employ \emph{data augmentation} (\da) \cite{Shorten2021,Feng2021}, which adds artificial training instances to a training set, in our case the training set of the end task. It is unclear, however, which \da methods improve most (if at all) the performance of pre-trained models per end-task 
\cite{Longpre2019,Longpre2020}.
Consequently, \citet{Feng2021} recommend exploring when \da is effective for large pre-trained models.

In this paper, we thoroughly examine the impact of \da in biomedical \qa, focusing on the factoid questions of the \bioasq challenge \cite{Tsatsaronis_et_al_bioasq} in an \mrc setting, i.e., we assume that relevant text passages, called \emph{snippets} in \bioasq, have already been identified. We first evaluate on \bioasq three indicative off-the-shelf pre-trained models, \distilbert \cite{DistilBERT}, \biobert \cite{biobert}, \albert \cite{albert}, already fine-tuned on \squad \cite{rajpurkar-2016-squad} or \squadtwo \cite{rajpurkar-etal-2018-squad-v2}, and we select 
\albert as our weak baseline. We also fine-tune \albert on \bioasq, on top of its \squad fine-tuning, to obtain a stronger baseline. We then obtain additional artificial training instances from an artificial cloze-style \mrc dataset, or via back-translation, information retrieval (\ir), word substitution based on \wtv or masked language modeling, question generation, or by extending the given passages with additional context.
\wtv-based word substitution, one of the simplest \da methods considered, improves test performance from 76.78\% precision-recall \auc 
(for \albert fine-tuned on \squad and \bioasq) to 84.99\%. 
Although we focus on biomedical \qa, our work should also be of interest in \qa for other specialized domains, e.g., legal \qa \cite{kien-etal-2020-answering,khazaeli-etal-2021-free,Zhang_2021}. 
Our work is the largest, in terms of \da methods considered, experimental study of \da for \qa (Section~\ref{sec:RelatedWork}).

Our main contributions are: (1) We present the largest (in terms of methods) 
experimental comparison of \da methods for \qa,
focusing on biomedical \qa, where obtaining training instances is particularly difficult and costly. (2) We show that \da can lead to very large performance gains, even when using pre-trained Transformers fine-tuned on large generic (\squad) and/or small domain-specific (\bioasq) end-task datasets, contributing to a broader discussion of if/when \da benefits pre-trained models. (3) We show that artificial cloze-style \mrc datasets are useful for \da. (4) We show that one of the simplest \da methods, \wtv-based word substitution, is also the best and is, therefore, recommended.
(5) We make our artificial training examples and code publicly available.\footnote{See \url{http://nlp.cs.aueb.gr/publications.html} for links to the code and data.}

\section{Experimental Setup}

\subsection{\bioasq Data in a \squad setting} \label{sec:data}

We experiment with data from \bioasq-8 (2021), Phase B, Task B \cite{Tsatsaronis_et_al_bioasq}, which contain English questions of biomedical experts.
Each question is accompanied by (i) the gold answer (often several alternative phrasings) and (ii) gold relevant passages, called \emph{snippets} (usually a single sentence each) from biomedical articles; the gold snippets contain the gold answer or other relevant information.
There are four question types: \textit{yes/no}, \textit{factoid},  \textit{list}, and questions requiring a \textit{summary}. We focus on factoid questions (e.g., ``\textit{Which gene is involved in Giant Axonal Neuropathy?}'').

We convert the \bioasq data to triples each containing a question, a single gold snippet, and the span of the gold answer in the snippet, much as in \squad \cite{rajpurkar-2016-squad}. If a question has multiple gold snippets, we produce equally many triples, discarding snippets that do not contain the gold answer. 
This conversion and considering only factoid questions allow us to use pre-trained Transformers already fine-tuned on \squad in a similar setting.\footnote{In the original \bioasq data, multiple snippets may be given for a particular question, the answer may be present in several of them, 
and identifying any answer span suffices.}
A disadvantage of the conversion is that our results are not directly comparable to those of \bioasq. The goal of our work, however, is to study the effect of different \da methods on a modern Transformer-based \qa baseline (and we show that fine-tuning it first on \squad helps), not to compete against \bioasq participants, who often use models tailored to the particular competition. 

From the $941$ factoid questions of the original \bioasq data, we obtained $3415$ question-snippet-answer triples.
We split these in training, development, test subsets ($2848$, $271$, $296$ triples, resp.), ensuring no question is in more than one subsets.

\subsection{Off-the-shelf Models} \label{sec:OffTheShelf}

As a starting point, we compared the performance of three publicly available pre-trained models that have already been fine-tuned for \mrc on \squad \cite{rajpurkar-2016-squad} or \squadtwo \cite{rajpurkar-etal-2018-squad-v2}.\footnote{We obtained the models from \url{https://huggingface.co/ktrapeznikov/albert-xlarge-v2-squad-v2}. We use the \textmc{xl} version of \albert. The other two models adopt the \textmc{bert-base} architecture; no \textmc{xl} variants were available.} At the time of our experiments, \albert-based models \cite{albert} were among the best on the \squad leaderboards; here, we use \albert fine-tuned on \squadtwo. We also considered \biobert \cite{biobert}, because it is pre-trained on a biomedical corpus; again, we use it fine-tuned on \squadtwo. The third model, \distilbert \cite{DistilBERT}, was chosen because of its much smaller size, which makes running experiments easier. This model is pretrained on a generic corpus, like the original \bert, and we use it fine-tuned on \squad. 
All three models are used here off-the-self, i.e., they are only evaluated, not trained in any way on \bioasq data. Throughout this work, we use the development subset of the \bioasq data to select models and configurations of \da methods, but in this experiment we use the union of the training and development subsets, since no \bioasq training is involved. 
\albert
is the best off-the-shelf model considered (Table~\ref{tab:PlugPlay}), hence we use it in all other experiments.\footnote{We discuss \prauc in Sections~\ref{sec:modifiedArchitecture} and \ref{sec:evaluationMeasures}.}

\begin{table}[ht]
\begin{center}
\resizebox{0.9\columnwidth}{!}{%
\begin{tabular}{cc}
\hline
\textbf{Model}  &
\textbf{\prauc (\bioasq train+dev)} 
\\
\hline
\hline
\distilbert (\squad) &
64.27 \\
\biobert (\squadtwo) &
69.22 \\
\textbf{\albert (\squadtwo)} &
\textbf{75.05} 
\\
\hline
\end{tabular}
}
\vspace*{-2mm}
\caption{\emph{Off-the-shelf} pre-trained models, fine-tuned for \mrc on \squad or \squadtwo. We report Precision-Recall \auc (\prauc, \%) on \bioasq training and development data, since no \bioasq training is involved. 
}
\vspace*{-6mm}
\label{tab:PlugPlay}
\end{center}
\end{table}

\begin{figure}[t]
\centering
\includegraphics[scale=0.35]{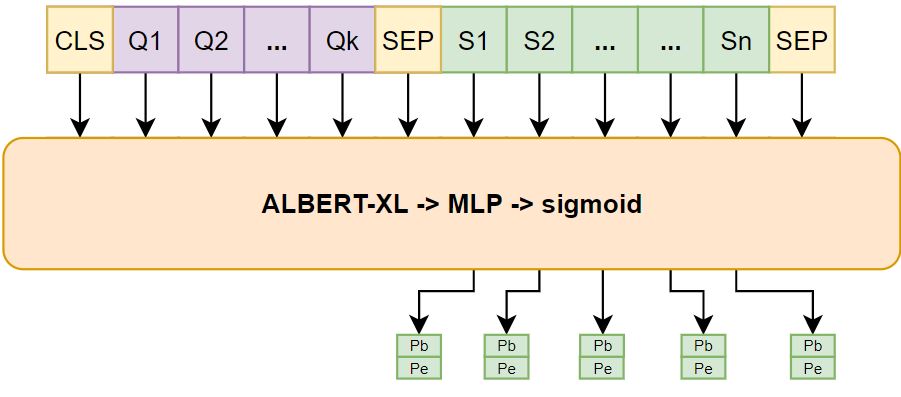}
\vspace*{-4mm}
\caption{The model used in all of the following experiments. \albertxl is fed with a question and snippet. Its contextualized embeddings are passed through an \mlp with \sigmoid activations that produces a begin ($P_b$) and end ($P_e$) probability per token of the snippet.
}
\vspace*{-6mm}
\label{fig:architecture}
\end{figure}

\subsection{Model Architecture Modifications} \label{sec:modifiedArchitecture}

The results of Table~\ref{tab:PlugPlay} were obtained by feeding the three off-the-shelf models with the concatenation of the question and snippet
of each question-snippet-answer \bioasq triple (training or development), without training of any kind. Following a typical \mrc architecture, each model was previously fine-tuned (by others) on \squad (or \squadtwo) with a shared
dense layer on top of each contextualized token embedding (of the snippet only) that the pre-trained model generates. The 
dense layer produces two logits per token, indicating the model's confidence that the token is the beginning or end of the answer, respectively. Two separate \softmax activations operate across all the begin and end logits, respectively, and the answer is the span (of the snippet) whose first and last tokens have the highest sum of begin and end probabilities (and the correct order).\footnote{In \squadtwo, additional layers decide if a question is answerable. We do not discuss them to save space.} The two \softmax activations presuppose that there is a single contiguous answer span in each snippet. This is true in \squad, but in our \bioasq data the (single) answer of a triple may consist of multiple non-contiguous spans of the triple's snippet (this happens in 583 out of 2,848 training instances). Hence, in the following experiments, where we further fine-tune \albert on \bioasq or artificial training data, we replace the two \softmax activations by two \sigmoid{s} that produce the begin and end probability per token of the snippet. Any token whose begin (or end) probability is above a threshold
$t$ is treated as the beginning (or end) of an answer span.
The \prauc evaluation measure (discussed below) aggregates results over different $t$ values. We also replace the dense layer on top of the contextualized token embeddings by a Multi-Layer Perceptron (\mlp) with a single hidden layer, which performed better on our development data. We use this single typical \mrc model architecture (Fig.\ref{fig:architecture}) in all the following experiments, since we aim to study the effect of several \da methods, not to propose new \mrc model architectures.

\subsection{Evaluation Measure} \label{sec:evaluationMeasures}

Given a development or test question-snippet-answer triple and a decision threshold $t$ (Section~\ref{sec:modifiedArchitecture}), we compute  precision and recall at the token level, i.e., we measure the ability of the model to identify the tokens of the answer. Precision is the number of correctly identified answer tokens, divided by the number of tokens in the model's answer. Recall is the number of correctly identified answer tokens, divided by the number of tokens in the correct answer. For different thresholds $t$, we obtain different precision-recall pairs for the same question-snippet-answer triple, which can be plotted as a precision-recall curve. We compute the trapezoidal area under the precision-recall curve (\prauc), and we 
then macro-average the \prauc scores over the test (or development) triples.\footnote{\prauc is similar to Mean Average Precision \cite{IRBook_MAP}, 
but obtains precision-recall points differently.}
\subsection{Baselines} \label{sec:baselines}

We use two baselines that do not involve \da: i) off-the-shelf \albert, pre-trained on a generic corpus, already fine-tuned on \squadtwo (last model of Table~\ref{tab:PlugPlay}); and ii) same as the first baseline, but further fine-tuned (on top of the fine-tuning on \squadtwo) on our \bioasq training data, with the modified architecture of Section~\ref{sec:modifiedArchitecture}.
Table~\ref{tab:baselines} shows that the second baseline is much stronger. Hence, we report performance gains with \da methods against the second baseline in subsequent sections.\footnote{We also experimented pre-trained \albert directly fine-tuned only on \bioasq, but the performance was much worse. 
}

\begin{table}[ht]
\begin{center}
\resizebox{\columnwidth}{!}{%
\begin{tabular}{ccc}
\hline
\textbf{Model} 
& \textbf{+train ex.}
& \textbf{\prauc (\bioasq dev)} 
\\
\hline
\hline
\albert (\squadtwo) & 0  & 80.25 \\
+\bioasq           & 2,848   & 89.57 \\
\hline
\end{tabular}
}
\vspace*{-2mm}
\caption{Performance of baselines on \bioasq dev.\ data. The first one is \albertxl fine-tuned on \squadtwo. The second one is also fine-tuned on \bioasq, with the modified architecture of Fig.~\ref{fig:architecture}. We also show the number of domain-specific (\bioasq) training examples.
}
\vspace*{-6mm}
\label{tab:baselines}
\end{center}
\end{table}

\section{Data Augmentation Methods}

There are two alternatives when using the artificial training instances that \da generates \citep{Yang_2019_DataAugRetrieval}. In our case, we always start with \albert, pre-trained on a generic corpus, and already fine-tuned on \squadtwo. In the first alternative, the model is then further fine-tuned on the artificial instances, and is then finally fine-tuned on the end-task data (\bioasq).
In the second alternative, the artificial and the end-task data are mixed, and the model is fine-tuned on the mixed data.
In each experiment below, we use the alternative (among the two) that leads to the best development \prauc.

\subsection{Artificial Cloze-style MRC Dataset}

For this augmentation method, we use \biomrc \citep{pappas-2020-biomrc}, the most recent and largest \emph{artificial} cloze-style biomedical \mrc dataset. \biomrc comes in two versions, \textmc{large} and \textmc{lite}, with 813k and 100k cloze-style questions, respectively. We use \biomrc \textmc{lite}.
 Each `question' is the title of a biomedical article, with an entity mentioned in the title hidden. Each question is accompanied by a passage (the abstract of the article), candidate answers (entities mentioned in the abstract), and the gold answer. 
From each passage we keep only the 
sentence containing the gold answer as the given snippet, and we generate a question-snippet-answer triple.\footnote{See the appendix for examples of all the \da methods.} If more than one sentences of the passage contain the gold answer, we create multiple triples, one for each sentence. We end up with approximately 142k artificial training triples.

\begin{table}[ht]
\begin{center}
\resizebox{\columnwidth}{!}{%
\begin{tabular}{ccc}
\hline
\textbf{\albert (\squadtwo)}
& \textbf{+train ex.}
& \textbf{\prauc (\bioasq dev)} 
\\
\hline
\hline
+\bioasq           & 2,848   & 89.57 \\ 
\hline
+\biomrc           & 2,848 & 78.66 \\
+\biomrc +\bioasq  & 5,696 & 91.57 \\
\hline
+\biomrc           & 10,000 & 91.20 \\
+\biomrc +\bioasq  & 12,848 & \textbf{93.15} \\
\hline
+\biomrc           & 30,000 & 90.57 \\
+\biomrc +\bioasq  & 32,848 & 92.19 \\
\hline
+\biomrc           & 50,000 & 91.19 \\
+\biomrc +\bioasq  & 52,848 & 91.51 \\
\hline
+\biomrc           & 100,000 & 90.93 \\
+\biomrc +\bioasq  & 102,848 & 92.39 \\
\hline
\end{tabular}
}
\vspace*{-2mm}
\caption{Adding training examples from an \emph{artificial cloze-style} \mrc dataset (\biomrc). The `+train ex.' column shows the number of domain-specific training examples (from \biomrc and/or \bioasq) used, on top of examples seen during fine-tuning on \squadtwo.
}
\vspace*{-4mm}
\label{tab:biomrc_res}
\end{center}
\end{table}

In Table~\ref{tab:biomrc_res}, the starting point is the weak baseline of Table~\ref{tab:baselines} (\albert fine-tuned on \squadtwo). We compare to the strong baseline (the second one of Table~\ref{tab:baselines}), which was further fine-tuned on \bioasq (+\bioasq). We show results when fine-tuning on \biomrc (+\biomrc) instead of \bioasq, and when fine-tuning on both \biomrc and \bioasq (+\biomrc +\bioasq), using 10k to 100k randomly sampled \biomrc examples. Interestingly, fine-tuning on 10k artificial \biomrc examples leads to better performance (91.20 dev.\ \prauc) than fine-tuning on \bioasq (89.57). The best performance (93.15) is obtained by fine-tuning on both \bioasq and 10k \biomrc examples. We attribute this improvement to the resemblance of \biomrc to \bioasq data. We see no benefit when adding more than 10k \biomrc examples, which may be an indication that the useful (for \bioasq) patterns that the model can learn from \biomrc are limited. 

\subsection{Back-translation} \label{sec:backtranslation}

Back translation (\btr) has been used for data augmentation in several \nlp tasks \cite{Feng2021,Shorten2021}.
The training examples are machine-translated from a source to a pivot 
language and back, obtaining paraphrases. We initially used French as the pivot language, then also Spanish and German, always translating from English to a pivot language and back with Google Translate.
For each question-snippet-answer training triple of \bioasq, we generate two new triples by back-translating either the question or the snippet. If a new triple is identical to the original one, we discard it. We obtained 4,901 new training examples pivoting only to French, and 15,593 when also pivoting to Spanish and German. 

\begin{table}[ht]
\begin{center}
\resizebox{\columnwidth}{!}{%
\begin{tabular}{ccc}
\hline
\textbf{\albert (\squadtwo)}
& \textbf{+train ex.}
& \textbf{\prauc (\bioasq dev)} 
\\
\hline
\hline \\
+\bioasq              & 2,848   & 89.57     \\ 
\hline
+\btr[\fr]          & 2,848 & 91.84 \\
+\btr[\fr] +\bioasq & 5,696 & \textbf{92.95} \\
\hline
+\btr[\fr]          & 4,901 & 89.80 \\ 
+\btr[\fr] +\bioasq & 7,749 & 91.44 \\ 
\hline 
+\btr[\fr,\es,\de]          & 2,848 & 89.80 \\
+\btr[\fr,\es,\de] +\bioasq & 5,696 & 89.99 \\
\hline 
+\btr[\fr,\es,\de] & 14,229 & 92.21\\
+\btr[\fr,\es,\de] +\bioasq & 17,077 & 92.21\\
\hline
\end{tabular}
}
\vspace*{-2mm}
\caption{Data augmentation via \emph{back-translation} (\btr), using one (\fr) or three (\fr, \es, \de) pivot languages. }
\vspace*{-4mm}
\label{tab:back_trans}
\end{center}
\end{table}

Table~\ref{tab:biomrc_res} shows that adding back-translations to the \bioasq training data increases development \prauc from 89.57 to 91.44 (or 92.66) with one (or three) pivot languages. Using back-translations with one pivot (+\btr [\fr]) instead of the original \bioasq data slightly surpasses the strong baseline (89.80 vs.\ 89.57); and with three pivots, using only back-translations (+\btr [\fr,\de,\es]) performs almost the same as adding the original \bioasq data too (92.52 vs.\ 92.66). These results show that \btr produces very good training instances and that further benefits may be possible with more pivots. Nevertheless simpler methods (e.g., \wtv-based word substitution, discussed below) offer larger gains with fewer artificial training instances.

\subsection{Information Retrieval} \label{sec:IR}

Data augmentation based on Information Retrieval (\ir) has been found promising in previous open-domain \qa work \cite{Yang_2019_DataAugRetrieval}.\footnote{\citet{Yang_2019_DataAugRetrieval} gained 2.7 to 9.7 F1 percentage points (pp.) in all four datasets they experimented with.}
Given a question and a gold answer, the question is used as a query to an \ir system.
Any retrieved document (or passage therein) that includes the gold answer is used to construct a new training example (with the same question and gold answer). We used the open data from the \pubmed Baseline Repository\footnote{\url{lhncbc.nlm.nih.gov/ii/information/MBR.html}} to create a pool of 22.3M biomedical documents. Each document is the concatenation of the title and abstract of a \pubmed article. We indexed all documents with an ElasticSearch retrieval engine\footnote{\url{https://www.elastic.co/}} and used the $500$ top ranked (by \bmtf) documents per question.
From the original 2,848 question-snippet-answer triples, only 289 more were generated, because in most of the retrieved documents no sentence included the entire answer (individual terms of the answer might be scattered in the document). We suspect that the biomedical experts of \bioasq create questions whose answers cannot be found in large numbers of documents (unlike common questions in open-domain \qa datasets), and the few relevant documents (and snippets) of each question have already been included in the \bioasq training data. 
Table~\ref{tab:ir_results} shows that \ir-based augmentation led to very minor gains in our case, because of the very few additional instances. 

\begin{table}[ht]
\begin{center}
\resizebox{\columnwidth}{!}{%
\begin{tabular}{ccc}
\hline
\textbf{\albert (\squadtwo)}
& \textbf{+train ex.}
& \textbf{\prauc (\bioasq dev)} 
\\
\hline
\hline
+\bioasq         & 2,848 & 89.57 \\ 
\hline
+\ir             & 289   & 80.30 \\ 
+\ir +\bioasq & 3,137 & \textbf{89.80} \\ 
\hline
\end{tabular}
}
\vspace*{-2mm}
\caption{Data augmentation via \emph{information retrieval} (\ir), using \pubmed titles and abstracts as documents.}
\vspace*{-5mm}
\label{tab:ir_results}
\end{center}
\end{table}

\subsection{Word Substitution} \label{sec:wordSubstitution}

These methods replace words of the original training examples by similar words (e.g., synonyms) from a thesaurus \cite{Jungiewicz_et_al_synonymsAug,Mahdi_et_al_OntoAug} or words with similar embeddings \cite{wang-yang-2015-thats}. More recent work uses large language models, pre-trained to predict masked tokens, which suggest replacements of randomly masked words of the original examples \cite{kobayashi_2018_contextual_augm,wu_et_al_2019_condit_bert_augm}. 

\subsubsection{\wtv-based Word Substitution} \label{sec:wordEmbeddingsSubstitution}

In this case, we use biomedical \wtv \cite{mikolov_w2v,brokos_et_al} embeddings.
Given a question-snippet-answer training instance, we consider all the word tokens of the snippet (excluding stop-words). For each token $w_i$ ($i=1, \dots, n$) of the snippet, we select the $k_i \leq K$ most similar words $w_j$ ($j=1, \dots, k_i$) of the vocabulary, using cosine similarity of word embeddings ($\vec{w}_i, \vec{w}_j$), that satisfy $\cos(\vec{w}_i, \vec{w}_j) \geq C$. We then produce $(k_1+1)(k_2+1)\dots(k_n+1) -1$ artificial training instances by replacing each token $w_i$ of the snippet by one of its $k_i$ most similar words (or itself), requiring at least one token of the original snippet to have been replaced. We then randomly sample 10k to 100k of the resulting instances and use them as additional training examples. We set $K=10$, $C=0.95$ based on preliminary experiments on development data. Adding 10k of the resulting artificial training examples to the original \bioasq examples leads to 95.60 development \prauc, outperforming the strong baseline (89.57) by six percentage points (Table~\ref{tab:w2v_lm_res}). Using only the 10k artificial examples, without any of the original examples, achieves almost identical performance (95.59), which suggests that the generated examples are of high quality. As when using artificial \mrc examples (Table~\ref{tab:biomrc_res}), adding more than 10k artificial instances provides no further benefit, probably because we end up adding too many minor variants of the same original examples.

\begin{table}[ht]
\begin{center}
\resizebox{\columnwidth}{!}{%
\begin{tabular}{ccc}
\hline
\textbf{\albert (\squadtwo)}
& \textbf{+train ex.}
& \textbf{\prauc (\bioasq dev)} 
\\
\hline
\hline
+ \bioasq         & 2,848 & 89.57 \\
\hline 
+\wtv           & 2,848 & 95.56 \\
+\wtv +\bioasq  & 5,696 & 95.27 \\
\hline
+\wtv  & 10,000  & 95.59 \\
+\wtv +\bioasq  & 12,848  & \textbf{95.60} \\
\hline
+\wtv  & 30,000  & 95.28 \\
+\wtv +\bioasq  & 32,848  & 95.20 \\
\hline
+\wtv  & 50,000  & 95.16 \\
+\wtv +\bioasq  & 52,848  & 95.13 \\
\hline
+\wtv  & 100,000 & 95.36 \\
+\wtv +\bioasq  & 102,848 & 95.22 \\
\hline
\end{tabular}
}
\vspace*{-2mm}
\caption{Data augmentation with \emph{\wtv-based word substitution}, using biomedical embeddings.}
\vspace*{-4mm}
\label{tab:w2v_lm_res}
\end{center}
\end{table}

The same \da mechanism could have been applied to questions instead of snippets.
In preliminary experiments, we employed an additional pre-trained natural language inference (\nli) model \cite{el-boukkouri-etal-2020-characterbert} as a \emph{consistency} mechanism to ensure the modified snippets followed from the original ones, but this also greatly reduced the number of artificial training instances we could generate. Performance was better without this mechanism, i.e., generating more artificial instances was better than generating fewer higher quality ones.

\subsubsection{Masked LM Word Substitution}\label{sec:lm_substitution}

Here we use \biolm \cite{lewis-etal-2020-pretrained} and specifically a \roberta{-\textmc{large}} model pre-trained on \pubmed, \pmc, and \mimicthree \cite{zhu2018clinical} with a new vocabulary extracted from \pubmed.\footnote{We did not use \biolm as an off-the-shelf \qa model (Section~\ref{sec:OffTheShelf}), because it was not available fine-tuned on \squad.} We use the same process as in \wtv word substitution, but each candidate replacement $w_j$ of an original word $w_i$ of the snippet must now satisfy $p(w_j) \geq P$ (instead of $\cos(\vec{w}_i, \vec{w}_j) \geq C$), where $p(w_j)$ is the probability assigned to $w_j$ by the pre-trained model; we also rank the candidate replacements $w_j$ of each $w_i$ by $p(w_j)$. We set 
$P=0.95$, based on preliminary experiments on development data.
Table~\ref{tab:bert_lm_res} shows that \biolm-based substitution is almost as good as \wtv-based substitution (94.45 vs.\ 95.60), but for \biolm the best performance is obtained with 50k artificial examples (compared to 10k for \wtv). This is probably due to the fact that \biolm suggests words that fit the particular context of the word being replaced and may, thus, suggest words with very different meanings that can be used in the particular context, adding noisy examples. By contrast, when using \wtv we compare more directly each original word with candidate replacements.\footnote{\wtv embeddings are not sensitive to the particular context of the snippet and rely exclusively on the (many more) contexts of each word encountered in the pre-training corpus.}

\begin{table}[ht]
\begin{center}
\resizebox{\columnwidth}{!}{%
\begin{tabular}{ccc}
\hline
\textbf{\albert (\squadtwo)}
& \textbf{+train ex.}
& \textbf{\prauc (\bioasq dev)} 
\\
\hline
\hline
+\bioasq             & 2,848 & 89.57 \\
\hline 
+\biolm            & 2,848 & 91.76 \\
+\biolm +\bioasq   & 5,696 & 92.37 \\
\hline
+\biolm            & 10,000  & 94.06 \\
+\biolm +\bioasq   & 12,848  & 94.06 \\
\hline
+\biolm            & 30,000  & 93.63 \\
+\biolm +\bioasq   & 32,848  & 93.75 \\
\hline
+\biolm            & 50,000  & 93.94 \\
+\biolm +\bioasq   & 52,848  & \textbf{94.45} \\
\hline
+\biolm            & 100,000 & 93.79 \\
+\biolm +\bioasq   & 102,848 & 93.84 \\
\hline
\hline
\end{tabular}
}
\vspace*{-2mm}
\caption{Data augmentation with \emph{word substitution} based on \emph{masked language modeling} using \biolm.}
\vspace*{-6mm}
\label{tab:bert_lm_res}
\end{center}
\end{table}

\subsection{Question Generation} \label{sec:questionGeneration}

Question generation (\qg) has been found an effective \da method in open-domain \mrc  \cite{alberti_2019_QuestGenRoundtripConsistency,chan_2019_RecBertQAGen,Enricoetal_2020_endtoend_qg}. The main reported benefit is that it 
increases the diversity of questions \cite{Qiuetal_19_rel_qg,sultan-etal-2020-importance}. Typically \qg 
models are fed with a snippet $s$, select an answer span 
$a$ of $s$, and generate a question $q$ answered by $a$. 
We take \tfivesp \citep{t5} fine-tuned for \qg on a modified version of \squad by \citet{Enricoetal_2020_endtoend_qg}\footnote{The \tfive\ \qg model we used is available at \url{https://github.com/patil-suraj/question_generation}.} and use it to generate alternative questions $q'$ and answer spans $a'$ from the snippets $s$ of the \bioasq $\left<q, s, a \right>$ training triples, producing artificial $\left<q', s, a' \right>$ triples. Multiple artificial triples can be generated from the same original one (the same $s$), but we require each $q'$  to be answered by a different answer span $a'$ to maximize the diversity of questions. We obtained 3,389 artificial triples from the 2,848 original ones this way. An alternative we explored is to select random snippets $s$ from random \pubmed abstracts, and use the \qg model to produce artificial $\left<q', s, a' \right>$ triples. The alternative approach can generate millions of artificial triples;
we generated up to 100k.

\begin{table}[ht]
\begin{center}
\resizebox{\columnwidth}{!}{%
\begin{tabular}{ccc}
\hline
\textbf{\albert (\squadtwo)}
& \textbf{+train ex.}
& \textbf{\prauc (\bioasq dev)}
\\
\hline
\hline
+\bioasq                & 2,848 & 89.57 \\
\hline
+\tfive@\bioasq             & 3,389 & 84.46 \\
+\tfive@\bioasq +\bioasq & 6,237 & 88.46 \\
\hline 
+\tfive@\pubmed          & 2,848 & 85.79 \\
+\tfive@\pubmed +\bioasq & 5,696 & 89.29 \\
\hline
+\tfive@\pubmed             & 10,000 & 87.30 \\ 
+\tfive@\pubmed +\bioasq & 12,848 & 89.34 \\
\hline
+\tfive@\pubmed             & 30,000 & 86.65 \\
+\tfive@\pubmed +\bioasq & 32,848 & 90.51 \\
\hline
+\tfive@\pubmed             & 50,000 & 87.30 \\
+\tfive@\pubmed +\bioasq & 52,848 & \textbf{90.69} \\
\hline
+\tfive@\pubmed             & 100,000 & 87.30 \\
+\tfive@\pubmed +\bioasq & 102,848 & 90.61 \\ 
\hline
\end{tabular}
}
\vspace*{-2mm}
\caption{Data augmentation via \emph{question generation} using \tfive. Questions are generated from the training snippets of \bioasq (\tfive@\bioasq) or from random snippets from random \pubmed abstracts (\tfive@\pubmed).}
\vspace*{-4mm}
\label{tab:quest_gen_res}
\end{center}
\end{table}

Table~\ref{tab:quest_gen_res} shows that adding to the \bioasq training data the artificial triples we obtained from \bioasq (+\tfive@\bioasq, \bioasq) is worse (88.46 vs.\ 89.57) than our strong baseline (+\bioasq). Fine-tuning only on the artificial triples (+\tfive@\bioasq) is much worse (84.46), i.e., the artificial triples are much less useful, despite being  more than the original ones. Adding artificial triples from \pubmed (+T5@\pubmed, \bioasq) performs only slightly better (90.69) than the strong baseline, when using 50k artificial triples, with no further benefit when using more.
A possible explanation for these poor results is the \tfive was fine-tuned for \qg on the open-domain \squad dataset. Thus, the generated questions are rather simplistic and not indicative of the specialized questions of \bioasq. Indeed, most of the generated questions are 
minor rephrases of 
the given snippet (e.g., subject replaced by `what').

\subsection{Adding Context} \label{sec:additionalContext}

In the original training  
question-snippet-answer 
$\left<q, s, a\right>$ triples,
$s$ is usually a single sentence. 
To help the \qa model learn to better distinguish relevant from irrelevant parts of the given snippet, we experimented with an additional \da method, where we find the original article that $s$ comes from and we expand $s$ with the $k_1$ (and $k_2$) sentences preceding (and following) it.\footnote{In \bioasq, each gold snippet is accompanied by the \pubmed id of the article it was extracted from.} For each original $\left<q, s, a\right>$ triple, we create multiple new $\left<q, s', a\right>$ artificial triples, for different values of $k_1 \geq 0$ and $k_2 \geq 0$, such that $k_1 + k_2 = K$.\footnote{Simply setting $k_1 = k_2$ would risk misguiding the model to always prefer the central sentence. We also experimented with \emph{random} $k_1$ (or $k_2$) sentences before (and after) $s$, but performance was much worse, possibly because the random sentences led to inferior context-aware token embeddings.} We experiment with $K=2$ (three new triples for each original one); then to obtain more artificial examples, we repeat with $K = 4$ (five new triples for each original).
To avoid truncation of the input examples, we remove all artificial examples that exceed $500$ characters in length.
For $K\in\{2,4\}$, we obtain a development \prauc score of 94.21 (Table~\ref{tab:cont_incr_res}), which is surpassed only by the the two embedding-based word substitution methods (Tables~\ref{tab:w2v_lm_res}--\ref{tab:bert_lm_res}). This \da method was introduced by \citet{previous_context_increasing}, who used it in \bioasq.\footnote{\citet{previous_context_increasing} reported an improvement in \bioasq's Lenient Accuracy by $2.49$ percentage points.}

\begin{table}[t]
\begin{center}
\resizebox{\columnwidth}{!}{%
\begin{tabular}{cccc}
\hline
\textbf{\albert (\squadtwo)}
& \textbf{+train ex.}
& \textbf{\prauc (\bioasq dev)}
\\
\hline
\hline
+\bioasq                  & 2,848 & 89.57 \\
\hline
+\context ($K=2$)               & 4,568 & 93.91 \\
+\context ($K=2$) +\bioasq   & 7,416 & 94.05 \\
\hline
+\context ($K \in \{2,4\}$)             & 6,428 & 94.20 \\
+\context ($K \in \{2,4\}$) +\bioasq & 9,276 & \textbf{94.21} \\
\hline
\end{tabular}
}
\vspace*{-2mm}
\caption{Data augmentation by \emph{adding context to the snippet} ($K=2$ or $K \in \{2, 4\}$ surrounding sentences).}
\vspace*{-8mm}
\label{tab:cont_incr_res}
\end{center}
\end{table}

\subsection{Final Results} \label{sec:testResults}

Table~\ref{tab:testResults} shows the performance of all the \da methods considered, on both development and test data. For each \da method, we use the configuration (from Tables~\ref{tab:biomrc_res}--\ref{tab:cont_incr_res}) with the best development score. The test scores are lower than the corresponding development ones, since several hyper-parameters (e.g., $K, C$ in the case of \wtv-based word substitution, number of training epochs) are tuned on  the development set. The test set also seems to be harder than the development one, since our weak baseline (\albert fine-tuned on \squadtwo with no further training) also performs worse on the test set (77.78 vs.\ 80.25). Nevertheless, the test scores confirm that \wtv-based word substitution is the best \da method considered, leading to a performance gain of 8.2 percentage points test \prauc compared to the strong baseline (84.99 vs.\ 77.78). The ranking of the other \da methods does not change when ranking by test score, instead of development score, with the only exception of adding context to the given passage (+\context), which is now slightly worse than adding instances from the artificial \biomrc dataset. Interestingly, all the \da methods, even the weakest \ir-based one, improve upon the test score of the strong baseline. 

\begin{table}[ht]
\begin{center}
\resizebox{\columnwidth}{!}{%
\begin{tabular}{cccc}
\hline
\textbf{Method}
& \textbf{+train ex.}
& \textbf{\prauc (dev)} 
& \textbf{\prauc (test)} 
\\
\hline
\hline
\albert (\squadtwo)                & 0     & 80.25 & 77.78 \\
+ \bioasq                             & 2,848 & 89.57 & 76.78 \\
\hline
+\wtv +\bioasq              & 12,848 & \textbf{95.60 (+6.03)} & \textbf{84.99 (+8.21)}\\
+\biolm +\bioasq            & 52,848 & 94.45 (+4.88) & 82.76 (+5.98)\\
+\context  +\bioasq         & 9,276 & 94.21 (+4.64) & 81.63 (+4.85)\\
+\biomrc +\bioasq           & 12,848 & 93.15 (+3.58) & 82.04 (+5.26)\\
+\btr +\bioasq              & 18,441 & 92.66 (+3.09) & 81.27 (+4.49) \\
+\tfive@\pubmed +\bioasq    & 52,848 & 90.69 (+1.12) & 80.26 (+3.48) \\
+\ir +\bioasq               & 3,137 & 89.80 (+0.23) & 78.66 (+1.88)\\
\hline
\end{tabular}
}
\vspace*{-3mm}
\caption{Performance of \da methods on \emph{development} and \emph{test} data, ordered by decreasing development score. For each \da method, we use the configuration (from Tables~\ref{tab:biomrc_res}--\ref{tab:cont_incr_res}) with the best development score.}
\vspace*{-6mm}
\label{tab:testResults}
\end{center}
\end{table}

\section{Related Work} \label{sec:RelatedWork}

\da is a key ingredient of success in deep learning for computer vision \cite{Shorten2019}. 
\da for \nlp has been explored less, but is an active research area \cite{Shorten2021,Feng2021}, with methods ranging from leveraging 
knowledge graphs  \cite{Moussallem_et_al_KnowGraphAug} to generating textual data from scratch
\cite{yang-etal-2020-generative,Bayer_et_al_GenerAug}.
The most common \nlp task in \da is text classification \cite{Bayer2021}. \citet{Feng2021} consider span-based \nlp tasks in specialized domains, which includes biomedical \mrc, among the most challenging for \da.

Word substitution is a simple and common \da approach in \nlp. In thesaurus-based substitution \cite{Jungiewicz_et_al_synonymsAug,Mahdi_et_al_OntoAug}, words 
are replaced by synonyms or closely related words (e.g., hypernyms).
Word embedding substitution \cite{wang-yang-2015-thats} replaces words by others nearby in a pre-trained vector space model (Section~\ref{sec:wordSubstitution}). Alternatively, a random word is removed,  inserted \cite{wei-zou-2019-eda,Miao_et_al}, or noised with spelling errors \cite{Belinkov_et_al_spelling_errors}. Sentences may also be re-ordered or removed \cite{Shen2020ASB,Chen2021HiddenCutSD}. Text generation has also been used in several \nlp tasks for adversarial augmentation \cite{Cheng2020AdvAugRA}, to paraphrase training examples \cite{ribeiro-etal-2018-semantically,cai-etal-2020-data,Xie2020UnsupervisedDA}, or generate new \cite{Anaby_Tavor_2020,kumar-etal-2020-data}. Back-translation \cite{Sennrich2016} is also widely used across \nlp tasks
\cite{Shorten2021,Feng2021}.

\da work for \qa in particular 
includes back-translation \cite{du_et_al_2019_hier_context_learning}, question generation \cite{zhang-bansal-2019-addressing,alberti_2019_QuestGenRoundtripConsistency,chan_2019_RecBertQAGen,Enricoetal_2020_endtoend_qg,yang-etal-2020-generative},
paraphrasing \cite{dong-etal-2017-learning,liu-etal-2020-tell}, and synonym replacement 
\cite{typogr_data_augm}, but not in a biomedical setting.
The \ir-based \da we used (Section~\ref{sec:IR}) follows 
\citet{Yang_2019_DataAugRetrieval}, 
who experimented in English and Chinese, but not in the biomedical domain. 
Expanding the passage with surrounding sentences (Section~\ref{sec:additionalContext}) follows 
\citet{previous_context_increasing}, who used this method in \bioasq.
\citet{dhingra-etal-2018-simple} create artificial cloze-style \mrc datasets and use them to pre-train neural \qa models (not Transformers), which are then fine-tuned on real training examples. By contrast, we use artificial \mrc datasets to fine-tune large pre-trained Transformers. 
All the above studies experimentally compare at most two \da methods; we compare seven.
Hence, our work is the largest (in terms of methods considered) experimental study of \da for \qa (and possibly \nlp).

\citet{Longpre2019} report that back-translation 
did not improve generalization in (non-biomedical) \qa experiments with fine-tuned pre-trained Transformers. \citet{Longpre2020} report that back-translation and Easy Data Augmentation \cite{Wei2019} are not always effective in text classification when fine-tuning pretrained Transformers, even with small end-task training sets. Consequently, \citet{Feng2021} recommend exploring when \da is effective for large pre-trained models. Our work contributes in this discussion by showing that \da can lead to very significant performance gains, even when using large pre-trained Transformers fine-tuned on large generic (\squad) and/or small domain-specific (\bioasq) end-task datasets. 

\section{Limitations and Future Work} \label{sec:limitations}

A limitation of our work is that we consider only \da in the \emph{input space}, i.e., the artificial instances are in textual form, like the original ones, as opposed to, e.g., interpolating feature vectors \cite{Chawla2002SMOTE,DeVries2017,Shorten2021}.
We also consider only \emph{offline} augmentation, i.e., the artificial instances are generated once, before training, as opposed to artificial instances generated anew for each training epoch. These two limitations, which are common in \da for \nlp, allow generating model-agnostic training instances once and reusing them across training epochs and different models. This greatly reduces computation costs and allows sharing the augmented datasets.
Online \da, however, exposes the model to many more synthetic data;
and feature space \da may act as layer-specific regularization.
One could also exploit ideas from active learning \cite{Ein-dor2020,Margatina2021} to select the most informative, diverse, and representative artificial training instances among those that \da generates. Small subsets of the \bioasq data could also be used to study the effect of \da in few-shot learning.

\bibliography{acl2020}

\begin{thebibliography}{87}
\expandafter\ifx\csname natexlab\endcsname\relax\def\natexlab#1{#1}\fi

\bibitem[{Abdollahi et~al.(2020)Abdollahi, Gao, Mei, Ghosh, and
  Li}]{Mahdi_et_al_OntoAug}
Mahdi Abdollahi, Xiaoying Gao, Yi~Mei, Shameek Ghosh, and Jinyan Li. 2020.
\newblock Ontology-guided data augmentation for medical document
  classification.
\newblock In \emph{Artificial Intelligence in Medicine}, pages 78--88, Cham.
  Springer International Publishing.

\bibitem[{Aghaebrahimian(2018)}]{aghaebrahimian-2018-linguistically}
Ahmad Aghaebrahimian. 2018.
\newblock Linguistically-based deep unstructured question answering.
\newblock In \emph{Proceedings of the 22nd Conference on Computational Natural
  Language Learning}, pages 433--443, Brussels, Belgium. Association for
  Computational Linguistics.

\bibitem[{Alberti et~al.(2019)Alberti, Andor, Pitler, Devlin, and
  Collins}]{alberti_2019_QuestGenRoundtripConsistency}
Chris Alberti, Daniel Andor, Emily Pitler, Jacob Devlin, and Michael Collins.
  2019.
\newblock Synthetic {QA} corpora generation with roundtrip consistency.
\newblock In \emph{Proceedings of the 57th Annual Meeting of the Association
  for Computational Linguistics}, pages 6168--6173, Florence, Italy.
  Association for Computational Linguistics.

\bibitem[{Anaby-Tavor et~al.(2020)Anaby-Tavor, Carmeli, Goldbraich, Kantor,
  Kour, Shlomov, Tepper, and Zwerdling}]{Anaby_Tavor_2020}
Ateret Anaby-Tavor, Boaz Carmeli, Esther Goldbraich, Amir Kantor, George Kour,
  Segev Shlomov, Naama Tepper, and Naama Zwerdling. 2020.
\newblock Do not have enough data? deep learning to the rescue!
\newblock \emph{Proceedings of the AAAI Conference on Artificial Intelligence},
  34(05):7383--7390.

\bibitem[{Bajgar et~al.(2016)Bajgar, Kadlec, and
  Kleindienst}]{Bajgar_et_al_BookTest}
Ondrej Bajgar, Rudolf Kadlec, and Jan Kleindienst. 2016.
\newblock {E}mbracing data abundance: {B}ook{T}est {D}ataset for {R}eading
  {C}omprehension.
\newblock \emph{CoRR}.

\bibitem[{Bayer et~al.(2021{\natexlab{a}})Bayer, Kaufhold, Buchhold, Keller,
  Dallmeyer, and Reuter}]{Bayer_et_al_GenerAug}
Markus Bayer, Marc-Andr{\'e} Kaufhold, Bj{\"o}rn Buchhold, Marcel Keller,
  J.~Dallmeyer, and Christian Reuter. 2021{\natexlab{a}}.
\newblock Data augmentation in natural language processing: A novel text
  generation approach for long and short text classifiers.
\newblock \emph{ArXiv}, abs/2103.14453.

\bibitem[{Bayer et~al.(2021{\natexlab{b}})Bayer, Kaufhold, and
  Reuter}]{Bayer2021}
Markus Bayer, Marc{-}Andr{\'{e}} Kaufhold, and Christian Reuter.
  2021{\natexlab{b}}.
\newblock \href {http://arxiv.org/abs/2107.03158} {A survey on data
  augmentation for text classification}.
\newblock \emph{CoRR}, abs/2107.03158.

\bibitem[{Belinkov and Bisk(2018)}]{Belinkov_et_al_spelling_errors}
Yonatan Belinkov and Yonatan Bisk. 2018.
\newblock Synthetic and natural noise both break neural machine translation.
\newblock \emph{ArXiv}, abs/1711.02173.

\bibitem[{Beltagy et~al.(2019)Beltagy, Lo, and Cohan}]{scibert}
Iz~Beltagy, Kyle Lo, and Arman Cohan. 2019.
\newblock \href {https://doi.org/10.18653/v1/D19-1371} {{S}ci{BERT}: A
  pretrained language model for scientific text}.
\newblock In \emph{Proceedings of the 2019 Conference on Empirical Methods in
  Natural Language Processing and the 9th International Joint Conference on
  Natural Language Processing (EMNLP-IJCNLP)}, pages 3615--3620, Hong Kong,
  China. Association for Computational Linguistics.

\bibitem[{Brokos et~al.(2018)Brokos, Liosis, McDonald, Pappas, and
  Androutsopoulos}]{brokos_et_al}
George Brokos, Polyvios Liosis, Ryan McDonald, Dimitris Pappas, and Ion
  Androutsopoulos. 2018.
\newblock {AUEB} at {B}io{ASQ} 6: {D}ocument and {S}nippet {R}etrieval.
\newblock In \emph{Proceedings of the 6th BioASQ Workshop}, pages 30--39,
  Brussels, Belgium.

\bibitem[{Cai et~al.(2020)Cai, Chen, Song, Zhang, Zhao, and
  Yin}]{cai-etal-2020-data}
Hengyi Cai, Hongshen Chen, Yonghao Song, Cheng Zhang, Xiaofang Zhao, and Dawei
  Yin. 2020.
\newblock Data manipulation: Towards effective instance learning for neural
  dialogue generation via learning to augment and reweight.
\newblock In \emph{Proceedings of the 58th Annual Meeting of the Association
  for Computational Linguistics}, pages 6334--6343, Online. Association for
  Computational Linguistics.

\bibitem[{Campos et~al.(2016)Campos, Nguyen, Rosenberg, Song, Gao, Tiwary,
  Majumder, Deng, and Mitra}]{Campos_2016_MSMARCO}
Daniel~Fernando Campos, T.~Nguyen, M.~Rosenberg, Xia Song, Jianfeng Gao,
  Saurabh Tiwary, Rangan Majumder, L.~Deng, and Bhaskar Mitra. 2016.
\newblock Ms marco: A human generated machine reading comprehension dataset.
\newblock \emph{ArXiv}, abs/1611.09268.

\bibitem[{Chalkidis et~al.(2020)Chalkidis, Fergadiotis, Malakasiotis, Aletras,
  and Androutsopoulos}]{legalbert}
Ilias Chalkidis, Manos Fergadiotis, Prodromos Malakasiotis, Nikolaos Aletras,
  and Ion Androutsopoulos. 2020.
\newblock \href {https://doi.org/10.18653/v1/2020.findings-emnlp.261}
  {{LEGAL}-{BERT}: The muppets straight out of law school}.
\newblock In \emph{Findings of the Association for Computational Linguistics:
  EMNLP 2020}, pages 2898--2904, Online. Association for Computational
  Linguistics.

\bibitem[{Chan and Fan(2019)}]{chan_2019_RecBertQAGen}
Ying-Hong Chan and Yao-Chung Fan. 2019.
\newblock A recurrent {BERT}-based model for question generation.
\newblock In \emph{Proceedings of the 2nd Workshop on Machine Reading for
  Question Answering}, pages 154--162, Hong Kong, China. Association for
  Computational Linguistics.

\bibitem[{Chawla et~al.(2002)Chawla, Bowyer, Hall, and
  Kegelmeyer}]{Chawla2002SMOTE}
Nitesh~V. Chawla, Kevin~W. Bowyer, Lawrence~O. Hall, and W.~Philip Kegelmeyer.
  2002.
\newblock Smote: Synthetic minority over-sampling technique.
\newblock \emph{Journal of Artificial Intelligence Research},
  16(1):321–--357.

\bibitem[{Chen et~al.(2016)Chen, Bolton, and
  Manning}]{chen-etal-2016-CNN-daily}
Danqi Chen, Jason Bolton, and Christopher~D. Manning. 2016.
\newblock A thorough examination of the {CNN}/{D}aily {M}ail reading
  comprehension task.
\newblock In \emph{Proceedings of the 54th Annual Meeting of the Association
  for Computational Linguistics (Volume 1: Long Papers)}, pages 2358--2367,
  Berlin, Germany. Association for Computational Linguistics.

\bibitem[{Chen et~al.(2017{\natexlab{a}})Chen, Fisch, Weston, and
  Bordes}]{dr_qa_2017}
Danqi Chen, Adam Fisch, Jason Weston, and Antoine Bordes. 2017{\natexlab{a}}.
\newblock \href {http://arxiv.org/abs/1704.00051} {Reading wikipedia to answer
  open-domain questions}.
\newblock \emph{CoRR}, abs/1704.00051.

\bibitem[{Chen et~al.(2017{\natexlab{b}})Chen, Fisch, Weston, and
  Bordes}]{chen_2017_DrQA}
Danqi Chen, Adam Fisch, Jason Weston, and Antoine Bordes. 2017{\natexlab{b}}.
\newblock Reading {Wikipedia} to answer open-domain questions.
\newblock In \emph{Association for Computational Linguistics (ACL)}.

\bibitem[{Chen et~al.(2021)Chen, Shen, Chen, and Yang}]{Chen2021HiddenCutSD}
Jiaao Chen, Dinghan Shen, Weizhu Chen, and Diyi Yang. 2021.
\newblock Hiddencut: Simple data augmentation for natural language
  understanding with better generalization.
\newblock \emph{ArXiv}, abs/2106.00149.

\bibitem[{Cheng et~al.(2020)Cheng, Jiang, Macherey, and
  Eisenstein}]{Cheng2020AdvAugRA}
Yong Cheng, Lu~Jiang, Wolfgang Macherey, and Jacob Eisenstein. 2020.
\newblock Advaug: Robust adversarial augmentation for neural machine
  translation.
\newblock \emph{ArXiv}, abs/2006.11834.

\bibitem[{Devlin et~al.(2019)Devlin, Chang, Lee, and Toutanova}]{bert}
Jacob Devlin, Ming-Wei Chang, Kenton Lee, and Kristina Toutanova. 2019.
\newblock \href {https://doi.org/10.18653/v1/N19-1423} {{BERT}: Pre-training of
  deep bidirectional transformers for language understanding}.
\newblock In \emph{Proceedings of the 2019 Conference of the North {A}merican
  Chapter of the Association for Computational Linguistics: Human Language
  Technologies, Volume 1 (Long and Short Papers)}, pages 4171--4186,
  Minneapolis, Minnesota.

\bibitem[{DeVries and Taylor(2017)}]{DeVries2017}
Terrance DeVries and Graham~W. Taylor. 2017.
\newblock \href {http://arxiv.org/abs/1702.05538} {Dataset augmentation in
  feature space}.

\bibitem[{Dhingra et~al.(2018)Dhingra, Danish, and
  Rajagopal}]{dhingra-etal-2018-simple}
Bhuwan Dhingra, Danish Danish, and Dheeraj Rajagopal. 2018.
\newblock \href {https://doi.org/10.18653/v1/N18-2092} {Simple and effective
  semi-supervised question answering}.
\newblock In \emph{Proceedings of the 2018 Conference of the North {A}merican
  Chapter of the Association for Computational Linguistics: Human Language
  Technologies, Volume 2 (Short Papers)}, pages 582--587, New Orleans,
  Louisiana.

\bibitem[{Dong et~al.(2017)Dong, Mallinson, Reddy, and
  Lapata}]{dong-etal-2017-learning}
Li~Dong, Jonathan Mallinson, Siva Reddy, and Mirella Lapata. 2017.
\newblock Learning to paraphrase for question answering.
\newblock In \emph{Proceedings of the 2017 Conference on Empirical Methods in
  Natural Language Processing}, pages 875--886, Copenhagen, Denmark.
  Association for Computational Linguistics.

\bibitem[{{Du} et~al.(2019){Du}, {Guo}, and
  {Zhao}}]{du_et_al_2019_hier_context_learning}
Y.~{Du}, W.~{Guo}, and Y.~{Zhao}. 2019.
\newblock Hierarchical question-aware context learning with augmented data for
  biomedical question answering.
\newblock In \emph{2019 IEEE International Conference on Bioinformatics and
  Biomedicine (BIBM)}, pages 370--375.

\bibitem[{Ein-Dor et~al.(2020)Ein-Dor, Halfon, Gera, Shnarch, Dankin, Choshen,
  Danilevsky, Aharonov, Katz, and Slonim}]{Ein-dor2020}
Liat Ein-Dor, Alon Halfon, Ariel Gera, Eyal Shnarch, Lena Dankin, Leshem
  Choshen, Marina Danilevsky, Ranit Aharonov, Yoav Katz, and Noam Slonim. 2020.
\newblock \href {https://doi.org/10.18653/v1/2020.emnlp-main.638} {{A}ctive
  {L}earning for {BERT}: {A}n {E}mpirical {S}tudy}.
\newblock In \emph{Proceedings of the 2020 Conference on Empirical Methods in
  Natural Language Processing (EMNLP)}, pages 7949--7962, Online.

\bibitem[{El~Boukkouri et~al.(2020)El~Boukkouri, Ferret, Lavergne, Noji,
  Zweigenbaum, and Tsujii}]{el-boukkouri-etal-2020-characterbert}
Hicham El~Boukkouri, Olivier Ferret, Thomas Lavergne, Hiroshi Noji, Pierre
  Zweigenbaum, and Jun{'}ichi Tsujii. 2020.
\newblock {C}haracter{BERT}: Reconciling {ELM}o and {BERT} for word-level
  open-vocabulary representations from characters.
\newblock In \emph{Proceedings of the 28th International Conference on
  Computational Linguistics}, pages 6903--6915, Barcelona, Spain (Online).
  International Committee on Computational Linguistics.

\bibitem[{Feng et~al.(2021)Feng, Gangal, Wei, Chandar, Vosoughi, Mitamura, and
  Hovy}]{Feng2021}
Steven~Y. Feng, Varun Gangal, Jason Wei, Sarath Chandar, Soroush Vosoughi,
  Teruko Mitamura, and Eduard Hovy. 2021.
\newblock \href {https://doi.org/10.18653/v1/2021.findings-acl.84} {A survey of
  data augmentation approaches for {NLP}}.
\newblock In \emph{Findings of the Association for Computational Linguistics:
  ACL-IJCNLP 2021}, pages 968--988, Online.

\bibitem[{Fu et~al.(2020)Fu, Qiu, Tang, Li, Yu, and Sun}]{Fu2020ASO}
Bin Fu, Yunqi Qiu, Chengguang Tang, Y.~Li, H.~Yu, and J.~Sun. 2020.
\newblock A survey on complex question answering over knowledge base: Recent
  advances and challenges.
\newblock \emph{ArXiv}, abs/2007.13069.

\bibitem[{He et~al.(2020)He, Liu, Gao, and Chen}]{deberta}
Pengcheng He, Xiaodong Liu, Jianfeng Gao, and Weizhu Chen. 2020.
\newblock \href {http://arxiv.org/abs/2006.03654} {Deberta: Decoding-enhanced
  bert with disentangled attention}.

\bibitem[{Hill et~al.(2016)Hill, Bordes, Chopra, and
  Weston}]{Hill_et_al_CBTest}
Felix Hill, Antoine Bordes, Sumit Chopra, and Jason Weston. 2016.
\newblock {T}he {G}oldilocks {P}rinciple: {R}eading {C}hildren's {B}ooks with
  {E}xplicit {M}emory {R}epresentations.
\newblock \emph{CoRR}.

\bibitem[{Jin et~al.(2019)Jin, Dhingra, Liu, Cohen, and Lu}]{jin-2019-pubmedqa}
Qiao Jin, Bhuwan Dhingra, Zhengping Liu, William Cohen, and Xinghua Lu. 2019.
\newblock {P}ub{M}ed{QA}: A dataset for biomedical research question answering.
\newblock In \emph{Proceedings of the 2019 Conference on Empirical Methods in
  Natural Language Processing and the 9th International Joint Conference on
  Natural Language Processing (EMNLP-IJCNLP)}, pages 2567--2577, Hong Kong,
  China. Association for Computational Linguistics.

\bibitem[{Joshi et~al.(2017)Joshi, Choi, Weld, and
  Zettlemoyer}]{joshi_2017_triviaqa}
Mandar Joshi, Eunsol Choi, Daniel Weld, and Luke Zettlemoyer. 2017.
\newblock {T}rivia{QA}: A large scale distantly supervised challenge dataset
  for reading comprehension.
\newblock In \emph{Proceedings of the 55th Annual Meeting of the Association
  for Computational Linguistics (Volume 1: Long Papers)}, pages 1601--1611,
  Vancouver, Canada. Association for Computational Linguistics.

\bibitem[{Jungiewicz and Smywinski-Pohl(2019)}]{Jungiewicz_et_al_synonymsAug}
Michał Jungiewicz and Aleksander Smywinski-Pohl. 2019.
\newblock Towards textual data augmentation for neural networks: synonyms and
  maximum loss.
\newblock \emph{Computer Science}, 20.

\bibitem[{Khazaeli et~al.(2021)Khazaeli, Punuru, Morris, Sharma, Staub, Cole,
  Chiu-Webster, and Sakalley}]{khazaeli-etal-2021-free}
Soha Khazaeli, Janardhana Punuru, Chad Morris, Sanjay Sharma, Bert Staub,
  Michael Cole, Sunny Chiu-Webster, and Dhruv Sakalley. 2021.
\newblock A free format legal question answering system.
\newblock In \emph{Proceedings of the Natural Legal Language Processing
  Workshop 2021}, pages 107--113, Punta Cana, Dominican Republic.

\bibitem[{Kien et~al.(2020)Kien, Nguyen, Bach, Tran, Nguyen, and
  Phuong}]{kien-etal-2020-answering}
Phi~Manh Kien, Ha-Thanh Nguyen, Ngo~Xuan Bach, Vu~Tran, Minh~Le Nguyen, and
  Tu~Minh Phuong. 2020.
\newblock Answering legal questions by learning neural attentive text
  representation.
\newblock In \emph{Proceedings of the 28th International Conference on
  Computational Linguistics}, pages 988--998, Barcelona, Spain (Online).
  International Committee on Computational Linguistics.

\bibitem[{Kobayashi(2018)}]{kobayashi_2018_contextual_augm}
Sosuke Kobayashi. 2018.
\newblock Contextual augmentation: Data augmentation by words with paradigmatic
  relations.
\newblock In \emph{Proceedings of the 2018 Conference of the North {A}merican
  Chapter of the Association for Computational Linguistics: Human Language
  Technologies, Volume 2 (Short Papers)}, pages 452--457.

\bibitem[{Kumar et~al.(2020)Kumar, Choudhary, and Cho}]{kumar-etal-2020-data}
Varun Kumar, Ashutosh Choudhary, and Eunah Cho. 2020.
\newblock Data augmentation using pre-trained transformer models.
\newblock In \emph{Proceedings of the 2nd Workshop on Life-long Learning for
  Spoken Language Systems}, pages 18--26, Suzhou, China. Association for
  Computational Linguistics.

\bibitem[{Kwiatkowski et~al.(2019)Kwiatkowski, Palomaki, Redfield, Collins,
  Parikh, Alberti, Epstein, Polosukhin, Devlin, Lee, Toutanova, Jones, Kelcey,
  Chang, Dai, Uszkoreit, Le, and Petrov}]{kwiatkowski-2019-natural}
Tom Kwiatkowski, Jennimaria Palomaki, Olivia Redfield, Michael Collins, Ankur
  Parikh, Chris Alberti, Danielle Epstein, Illia Polosukhin, Jacob Devlin,
  Kenton Lee, Kristina Toutanova, Llion Jones, Matthew Kelcey, Ming-Wei Chang,
  Andrew~M. Dai, Jakob Uszkoreit, Quoc Le, and Slav Petrov. 2019.
\newblock Natural questions: A benchmark for question answering research.
\newblock \emph{Transactions of the Association for Computational Linguistics},
  7:452--466.

\bibitem[{Lai et~al.(2017)Lai, Xie, Liu, Yang, and Hovy}]{lai-2017-race}
Guokun Lai, Qizhe Xie, Hanxiao Liu, Yiming Yang, and Eduard Hovy. 2017.
\newblock {RACE}: Large-scale {R}e{A}ding comprehension dataset from
  examinations.
\newblock In \emph{Proceedings of the 2017 Conference on Empirical Methods in
  Natural Language Processing}, pages 785--794, Copenhagen, Denmark.
  Association for Computational Linguistics.

\bibitem[{Lan et~al.(2019)Lan, Chen, Goodman, Gimpel, Sharma, and
  Soricut}]{albert}
Zhenzhong Lan, Mingda Chen, Sebastian Goodman, Kevin Gimpel, Piyush Sharma, and
  Radu Soricut. 2019.
\newblock {ALBERT:} {A} lite {BERT} for self-supervised learning of language
  representations.
\newblock \emph{CoRR}, abs/1909.11942.

\bibitem[{Lee et~al.(2019)Lee, Yoon, Kim, Kim, Kim, So, and Kang}]{biobert}
Jinhyuk Lee, Wonjin Yoon, Sungdong Kim, Donghyeon Kim, Sunkyu Kim, Chan~Ho So,
  and Jaewoo Kang. 2019.
\newblock {BioBERT: a pre-trained biomedical language representation model for
  biomedical text mining}.
\newblock \emph{Bioinformatics}, 36(4):1234--1240.

\bibitem[{Lewis et~al.(2020)Lewis, Ott, Du, and
  Stoyanov}]{lewis-etal-2020-pretrained}
Patrick Lewis, Myle Ott, Jingfei Du, and Veselin Stoyanov. 2020.
\newblock Pretrained language models for biomedical and clinical tasks:
  Understanding and extending the state-of-the-art.
\newblock In \emph{Proceedings of the 3rd Clinical Natural Language Processing
  Workshop}, pages 146--157, Online. Association for Computational Linguistics.

\bibitem[{Liu et~al.(2020)Liu, Gong, Fu, Yan, Chen, Lv, Duan, and
  Zhou}]{liu-etal-2020-tell}
Dayiheng Liu, Yeyun Gong, Jie Fu, Yu~Yan, Jiusheng Chen, Jiancheng Lv, Nan
  Duan, and Ming Zhou. 2020.
\newblock Tell me how to ask again: Question data augmentation with
  controllable rewriting in continuous space.
\newblock In \emph{Proceedings of the 2020 Conference on Empirical Methods in
  Natural Language Processing (EMNLP)}, pages 5798--5810, Online. Association
  for Computational Linguistics.

\bibitem[{Liu et~al.(2019)Liu, Ott, Goyal, Du, Joshi, Chen, Levy, Lewis,
  Zettlemoyer, and Stoyanov}]{roberta}
Yinhan Liu, Myle Ott, Naman Goyal, Jingfei Du, Mandar Joshi, Danqi Chen, Omer
  Levy, Mike Lewis, Luke Zettlemoyer, and Veselin Stoyanov. 2019.
\newblock \href {http://arxiv.org/abs/1907.11692} {Roberta: A robustly
  optimized bert pretraining approach}.

\bibitem[{Longpre et~al.(2019)Longpre, Lu, Tu, and DuBois}]{Longpre2019}
Shayne Longpre, Yi~Lu, Zhucheng Tu, and Chris DuBois. 2019.
\newblock \href {http://arxiv.org/abs/1912.02145} {An exploration of data
  augmentation and sampling techniques for domain-agnostic question answering}.
\newblock \emph{CoRR}, abs/1912.02145.

\bibitem[{Longpre et~al.(2020)Longpre, Wang, and DuBois}]{Longpre2020}
Shayne Longpre, Yu~Wang, and Chris DuBois. 2020.
\newblock \href {https://doi.org/10.18653/v1/2020.findings-emnlp.394} {How
  effective is task-agnostic data augmentation for pretrained transformers?}
\newblock In \emph{Findings of the Association for Computational Linguistics:
  EMNLP 2020}, pages 4401--4411, Online.

\bibitem[{Lopez et~al.(2020)Lopez, Cruz, Cruz, and
  Cheng}]{Enricoetal_2020_endtoend_qg}
Luis~Enrico Lopez, Diane~Kathryn Cruz, Jan Christian~Blaise Cruz, and Charibeth
  Cheng. 2020.
\newblock Transformer-based end-to-end question generation.
\newblock \emph{CoRR}, abs/2005.01107.

\bibitem[{Luo et~al.(2018)Luo, Lin, Luo, and Zhu}]{luo-etal-2018-knowledge}
Kangqi Luo, Fengli Lin, Xusheng Luo, and Kenny Zhu. 2018.
\newblock Knowledge base question answering via encoding of complex query
  graphs.
\newblock In \emph{Proceedings of the 2018 Conference on Empirical Methods in
  Natural Language Processing}, pages 2185--2194, Brussels, Belgium.
  Association for Computational Linguistics.

\bibitem[{Manning et~al.(2008)Manning, Raghavan, and Sch\"{u}tze}]{IRBook_MAP}
Christopher~D. Manning, Prabhakar Raghavan, and Hinrich Sch\"{u}tze. 2008.
\newblock \emph{Introduction to Information Retrieval}.
\newblock Cambridge University Press.

\bibitem[{Margatina et~al.(2021)Margatina, Vernikos, Barrault, and
  Aletras}]{Margatina2021}
Katerina Margatina, Giorgos Vernikos, Lo{\"\i}c Barrault, and Nikolaos Aletras.
  2021.
\newblock \href {https://aclanthology.org/2021.emnlp-main.51} {Active learning
  by acquiring contrastive examples}.
\newblock In \emph{Proceedings of the 2021 Conference on Empirical Methods in
  Natural Language Processing}, pages 650--663, Online and Punta Cana,
  Dominican Republic.

\bibitem[{Miao et~al.(2020)Miao, Li, Wang, and Tan}]{Miao_et_al}
Zhengjie Miao, Yuliang Li, Xiaolan Wang, and Wang-Chiew Tan. 2020.
\newblock Snippext: Semi-supervised opinion mining with augmented data.
\newblock In \emph{Proceedings of The Web Conference 2020}, page 617–628, New
  York, NY, USA. Association for Computing Machinery.

\bibitem[{Mikolov et~al.(2013)Mikolov, Chen, Corrado, and Dean}]{mikolov_w2v}
Tom{\'{a}}s Mikolov, Kai Chen, Greg Corrado, and Jeffrey Dean. 2013.
\newblock Efficient estimation of word representations in vector space.
\newblock In \emph{1st International Conference on Learning Representations,
  {ICLR} 2013, Scottsdale, Arizona, USA, May 2-4, 2013, Workshop Track
  Proceedings}.

\bibitem[{M{\"o}ller et~al.(2020)M{\"o}ller, Reina, Jayakumar, and
  Pietsch}]{moller-2020-COVID-QA}
Timo M{\"o}ller, Anthony Reina, Raghavan Jayakumar, and Malte Pietsch. 2020.
\newblock {COVID-QA}: A question answering dataset for {COVID}-19.
\newblock In \emph{Proceedings of the 1st Workshop on {NLP} for {COVID-19} at
  {ACL} 2020}, Online. Association for Computational Linguistics.

\bibitem[{Moussallem et~al.(2019)Moussallem, Arcan, Ngomo, and
  Buitelaar}]{Moussallem_et_al_KnowGraphAug}
Diego Moussallem, Mihael Arcan, Axel{-}Cyrille~Ngonga Ngomo, and Paul
  Buitelaar. 2019.
\newblock Augmenting neural machine translation with knowledge graphs.
\newblock \emph{CoRR}, abs/1902.08816.

\bibitem[{{Nugraha} and {Suyanto}(2019)}]{typogr_data_augm}
H.~S. {Nugraha} and S.~{Suyanto}. 2019.
\newblock Typographic-based data augmentation to improve a question retrieval
  in short dialogue system.
\newblock In \emph{2019 International Seminar on Research of Information
  Technology and Intelligent Systems (ISRITI)}, pages 44--49.

\bibitem[{Pappas and
  Androutsopoulos(2021)}]{pappas-androutsopoulos-2021-neural}
Dimitris Pappas and Ion Androutsopoulos. 2021.
\newblock A neural model for joint document and snippet ranking in question
  answering for large document collections.
\newblock In \emph{Proceedings of the 59th Annual Meeting of the Association
  for Computational Linguistics and the 11th International Joint Conference on
  Natural Language Processing (Volume 1: Long Papers)}, pages 3896--3907,
  Online. Association for Computational Linguistics.

\bibitem[{Pappas et~al.(2018)Pappas, Androutsopoulos, and
  Papageorgiou}]{pappas-2018-bioread}
Dimitris Pappas, Ion Androutsopoulos, and Haris Papageorgiou. 2018.
\newblock {B}io{R}ead: A new dataset for biomedical reading comprehension.
\newblock In \emph{Proceedings of the Eleventh International Conference on
  Language Resources and Evaluation ({LREC} 2018)}, Miyazaki, Japan. European
  Language Resources Association (ELRA).

\bibitem[{Pappas et~al.(2020)Pappas, Stavropoulos, Androutsopoulos, and
  McDonald}]{pappas-2020-biomrc}
Dimitris Pappas, Petros Stavropoulos, Ion Androutsopoulos, and Ryan McDonald.
  2020.
\newblock {B}io{MRC}: A dataset for biomedical machine reading comprehension.
\newblock In \emph{Proceedings of the 19th SIGBioMed Workshop on Biomedical
  Language Processing}, pages 140--149, Online. Association for Computational
  Linguistics.

\bibitem[{Qiu and Xiong(2019)}]{Qiuetal_19_rel_qg}
Jiazuo Qiu and Deyi Xiong. 2019.
\newblock Generating highly relevant questions.
\newblock \emph{CoRR}, abs/1910.03401.

\bibitem[{Raffel et~al.(2020)Raffel, Shazeer, Roberts, Lee, Narang, Matena,
  Zhou, Li, and Liu}]{t5}
Colin Raffel, Noam Shazeer, Adam Roberts, Katherine Lee, Sharan Narang, Michael
  Matena, Yanqi Zhou, Wei Li, and Peter~J. Liu. 2020.
\newblock \href {http://jmlr.org/papers/v21/20-074.html} {Exploring the limits
  of transfer learning with a unified text-to-text transformer}.
\newblock \emph{Journal of Machine Learning Research}, 21(140):1--67.

\bibitem[{Rajpurkar et~al.(2018)Rajpurkar, Jia, and
  Liang}]{rajpurkar-etal-2018-squad-v2}
Pranav Rajpurkar, Robin Jia, and Percy Liang. 2018.
\newblock Know what you don{'}t know: Unanswerable questions for {SQ}u{AD}.
\newblock In \emph{Proceedings of the 56th Annual Meeting of the Association
  for Computational Linguistics (Volume 2: Short Papers)}, pages 784--789,
  Melbourne, Australia. Association for Computational Linguistics.

\bibitem[{Rajpurkar et~al.(2016)Rajpurkar, Zhang, Lopyrev, and
  Liang}]{rajpurkar-2016-squad}
Pranav Rajpurkar, Jian Zhang, Konstantin Lopyrev, and Percy Liang. 2016.
\newblock {SQ}u{AD}: 100,000+ questions for machine comprehension of text.
\newblock In \emph{Proceedings of the 2016 Conference on Empirical Methods in
  Natural Language Processing}, pages 2383--2392, Austin, Texas. Association
  for Computational Linguistics.

\bibitem[{Reddy et~al.(2019)Reddy, Chen, and Manning}]{reddy-2019-coqa}
Siva Reddy, Danqi Chen, and Christopher~D. Manning. 2019.
\newblock {C}o{QA}: A conversational question answering challenge.
\newblock \emph{Transactions of the Association for Computational Linguistics},
  7:249--266.

\bibitem[{Ribeiro et~al.(2018)Ribeiro, Singh, and
  Guestrin}]{ribeiro-etal-2018-semantically}
Marco~Tulio Ribeiro, Sameer Singh, and Carlos Guestrin. 2018.
\newblock Semantically equivalent adversarial rules for debugging {NLP} models.
\newblock In \emph{Proceedings of the 56th Annual Meeting of the Association
  for Computational Linguistics (Volume 1: Long Papers)}, pages 856--865,
  Melbourne, Australia. Association for Computational Linguistics.

\bibitem[{Sanh et~al.(2019)Sanh, Debut, Chaumond, and Wolf}]{DistilBERT}
Victor Sanh, Lysandre Debut, Julien Chaumond, and Thomas Wolf. 2019.
\newblock Distilbert, a distilled version of {BERT:} smaller, faster, cheaper
  and lighter.
\newblock \emph{CoRR}, abs/1910.01108.

\bibitem[{Sennrich et~al.(2016)Sennrich, Haddow, and Birch}]{Sennrich2016}
Rico Sennrich, Barry Haddow, and Alexandra Birch. 2016.
\newblock \href {https://doi.org/10.18653/v1/P16-1009} {Improving neural
  machine translation models with monolingual data}.
\newblock In \emph{Proceedings of the 54th Annual Meeting of the Association
  for Computational Linguistics (Volume 1: Long Papers)}, pages 86--96, Berlin,
  Germany.

\bibitem[{Shen et~al.(2020)Shen, Zheng, Shen, Qu, and Chen}]{Shen2020ASB}
Dinghan Shen, Ming Zheng, Yelong Shen, Yanru Qu, and Weizhu Chen. 2020.
\newblock A simple but tough-to-beat data augmentation approach for natural
  language understanding and generation.
\newblock \emph{ArXiv}, abs/2009.13818.

\bibitem[{Shorten and Khoshgoftaar(2019)}]{Shorten2019}
Connor Shorten and Taghi~M. Khoshgoftaar. 2019.
\newblock A survey on image data augmentation for deep learning.
\newblock \emph{Journal of Big Data}, 6(60).

\bibitem[{Shorten et~al.(2021)Shorten, Khoshgoftaar, and Furht}]{Shorten2021}
Connor Shorten, Taghi~M. Khoshgoftaar, and Borko Furht. 2021.
\newblock Text data augmentation for deep learning.
\newblock \emph{Journal of Big Data}, 8(101).

\bibitem[{Sultan et~al.(2020)Sultan, Chandel, Fernandez~Astudillo, and
  Castelli}]{sultan-etal-2020-importance}
Md~Arafat Sultan, Shubham Chandel, Ram{\'o}n Fernandez~Astudillo, and Vittorio
  Castelli. 2020.
\newblock On the importance of diversity in question generation for {QA}.
\newblock In \emph{Proceedings of the 58th Annual Meeting of the Association
  for Computational Linguistics}, pages 5651--5656, Online. Association for
  Computational Linguistics.

\bibitem[{Tsatsaronis et~al.(2015)Tsatsaronis, Balikas, Malakasiotis, Partalas,
  Zschunke, Alvers, Weissenborn, Krithara, Petridis, Polychronopoulos,
  Almirantis, Pavlopoulos, Baskiotis, Gallinari, Artieres, Ngonga, Heino,
  Gaussier, Barrio-Alvers, Schroeder, Androutsopoulos, and
  Paliouras}]{Tsatsaronis_et_al_bioasq}
G.~Tsatsaronis, G.~Balikas, P.~Malakasiotis, I.~Partalas, M.~Zschunke, M.R.
  Alvers, D.~Weissenborn, A.~Krithara, S.~Petridis, D.~Polychronopoulos,
  Y.~Almirantis, J.~Pavlopoulos, N.~Baskiotis, P.~Gallinari, T.~Artieres,
  A.~Ngonga, N.~Heino, E.~Gaussier, L.~Barrio-Alvers, M.~Schroeder,
  I.~Androutsopoulos, and G.~Paliouras. 2015.
\newblock {A}n {O}verview of the {B}io{ASQ} {L}arge-{S}cale {B}iomedical
  {S}emantic {I}ndexing and {Q}uestion {A}nswering {C}ompetition.
\newblock \emph{BMC Bioinformatics}, 16(138).

\bibitem[{Wang et~al.(2020)Wang, Yao, Zhang, Xu, and Wang}]{Wang_2020_ReCO}
Bingning Wang, Ting Yao, Qi~Zhang, Jingfang Xu, and Xiaochuan Wang. 2020.
\newblock Reco: A large scale chinese reading comprehension dataset on opinion.
\newblock \emph{Proceedings of the AAAI Conference on Artificial Intelligence},
  34(05):9146--9153.

\bibitem[{Wang and Yang(2015)}]{wang-yang-2015-thats}
William~Yang Wang and Diyi Yang. 2015.
\newblock That{'}s so annoying!!!: A lexical and frame-semantic embedding based
  data augmentation approach to automatic categorization of annoying behaviors
  using {\#}petpeeve tweets.
\newblock In \emph{Proceedings of the 2015 Conference on Empirical Methods in
  Natural Language Processing}, pages 2557--2563, Lisbon, Portugal. Association
  for Computational Linguistics.

\bibitem[{Wei and Zou(2019{\natexlab{a}})}]{wei-zou-2019-eda}
Jason Wei and Kai Zou. 2019{\natexlab{a}}.
\newblock {EDA}: Easy data augmentation techniques for boosting performance on
  text classification tasks.
\newblock In \emph{Proceedings of the 2019 Conference on Empirical Methods in
  Natural Language Processing and the 9th International Joint Conference on
  Natural Language Processing (EMNLP-IJCNLP)}, pages 6382--6388, Hong Kong,
  China. Association for Computational Linguistics.

\bibitem[{Wei and Zou(2019{\natexlab{b}})}]{Wei2019}
Jason Wei and Kai Zou. 2019{\natexlab{b}}.
\newblock \href {https://doi.org/10.18653/v1/D19-1670} {{EDA}: Easy data
  augmentation techniques for boosting performance on text classification
  tasks}.
\newblock In \emph{Proceedings of the 2019 Conference on Empirical Methods in
  Natural Language Processing and the 9th International Joint Conference on
  Natural Language Processing (EMNLP-IJCNLP)}, pages 6382--6388, Hong Kong,
  China.

\bibitem[{Wu et~al.(2019)Wu, Lv, Zang, Han, and
  Hu}]{wu_et_al_2019_condit_bert_augm}
Xing Wu, Shangwen Lv, Liangjun Zang, Jizhong Han, and Songlin Hu. 2019.
\newblock Conditional bert contextual augmentation.
\newblock In \emph{International Conference on Computational Science 2019},
  pages 84--95, Cham.

\bibitem[{Xie et~al.(2020)Xie, Dai, Hovy, Luong, and
  Le}]{Xie2020UnsupervisedDA}
Qizhe Xie, Zihang Dai, E.~Hovy, Minh-Thang Luong, and Quoc~V. Le. 2020.
\newblock Unsupervised data augmentation for consistency training.
\newblock \emph{arXiv: Learning}.

\bibitem[{Yadati et~al.(2021)Yadati, R~S, S, K~M, and
  G}]{yadati-etal-2021-knowledge}
Naganand Yadati, Dayanidhi R~S, Vaishnavi S, Indira K~M, and Srinidhi G. 2021.
\newblock Knowledge base question answering through recursive hypergraphs.
\newblock In \emph{Proceedings of the 16th Conference of the European Chapter
  of the Association for Computational Linguistics: Main Volume}, pages
  448--454, Online. Association for Computational Linguistics.

\bibitem[{Yang et~al.(2019)Yang, Xie, Tan, Xiong, Li, and
  Lin}]{Yang_2019_DataAugRetrieval}
Wei Yang, Yuqing Xie, Luchen Tan, Kun Xiong, M.~Li, and Jimmy Lin. 2019.
\newblock Data augmentation for bert fine-tuning in open-domain question
  answering.
\newblock \emph{ArXiv}, abs/1904.06652.

\bibitem[{Yang et~al.(2015)Yang, Yih, and Meek}]{yang-2015-wikiqa}
Yi~Yang, Wen-tau Yih, and Christopher Meek. 2015.
\newblock {W}iki{QA}: A challenge dataset for open-domain question answering.
\newblock In \emph{Proceedings of the 2015 Conference on Empirical Methods in
  Natural Language Processing}, pages 2013--2018, Lisbon, Portugal. Association
  for Computational Linguistics.

\bibitem[{Yang et~al.(2020)Yang, Malaviya, Fernandez, Swayamdipta, Le~Bras,
  Wang, Bhagavatula, Choi, and Downey}]{yang-etal-2020-generative}
Yiben Yang, Chaitanya Malaviya, Jared Fernandez, Swabha Swayamdipta, Ronan
  Le~Bras, Ji-Ping Wang, Chandra Bhagavatula, Yejin Choi, and Doug Downey.
  2020.
\newblock Generative data augmentation for commonsense reasoning.
\newblock In \emph{Findings of the Association for Computational Linguistics:
  EMNLP 2020}, pages 1008--1025, Online. Association for Computational
  Linguistics.

\bibitem[{Yang and Choi(2019)}]{yang-choi-2019-friendsqa}
Zhengzhe Yang and Jinho~D. Choi. 2019.
\newblock {F}riends{QA}: Open-domain question answering on {TV} show
  transcripts.
\newblock In \emph{Proceedings of the 20th Annual SIGdial Meeting on Discourse
  and Dialogue}, pages 188--197, Stockholm, Sweden. Association for
  Computational Linguistics.

\bibitem[{Yoon et~al.(2020)Yoon, Lee, Kim, Jeong, and
  Kang}]{previous_context_increasing}
Wonjin Yoon, Jinhyuk Lee, Donghyeon Kim, Minbyul Jeong, and Jaewoo Kang. 2020.
\newblock Pre-trained language model for biomedical question answering.
\newblock In \emph{Machine Learning and Knowledge Discovery in Databases},
  pages 727--740, Cham. Springer International Publishing.

\bibitem[{Zhang and Xing(2021)}]{Zhang_2021}
Na-Na Zhang and Yinan Xing. 2021.
\newblock Questions and answers on legal texts based on {BERT}-{BiGRU}.
\newblock \emph{Journal of Physics: Conference Series}, 1828(1):012035.

\bibitem[{Zhang and Bansal(2019)}]{zhang-bansal-2019-addressing}
Shiyue Zhang and Mohit Bansal. 2019.
\newblock Addressing semantic drift in question generation for semi-supervised
  question answering.
\newblock In \emph{Proceedings of the 2019 Conference on Empirical Methods in
  Natural Language Processing and the 9th International Joint Conference on
  Natural Language Processing (EMNLP-IJCNLP)}, pages 2495--2509, Hong Kong,
  China. Association for Computational Linguistics.

\bibitem[{Zhu et~al.(2018)Zhu, Paschalidis, and Tahmasebi}]{zhu2018clinical}
Henghui Zhu, Ioannis~Ch Paschalidis, and Amir Tahmasebi. 2018.
\newblock Clinical concept extraction with contextual word embedding.
\newblock \emph{arXiv preprint arXiv:1810.10566}.

\end{thebibliography}
\bibliographystyle{acl_natbib}

\setcounter{table}{0}
\renewcommand{\thetable}{\Alph{section}\arabic{table}}
\appendix

\section*{Appendix}

\section{Combining  Augmentation Methods} \label{sec:ensemble}

We also tried to combine \da methods. In Table~\ref{tab:ensemble_res}, we incrementally add to the training set of the strong baseline (\albert fine-tuned on \squadtwo, then \bioasq) artificial training examples obtained from \wtv-based word substitution, then (additionally) training examples obtained by expanding the context of the given passage etc. We started with the artificial examples of the \wtv-based method, which had the best development score, skipped the other (\biolm-based) word substitution method, then continued with examples from \biomrc and back-translation, which were the next best in terms of development score. Unfortunately, there was no significant gain, compared to using only the \wtv-based method, which suggests that the \da methods we consider are not complementary. An alternative approach would be to stack \da methods, instead of accumulating their training examples. For example, one could apply the \wtv method to artificial examples produced by increasing the context of the given passages. We leave this for future work.

\begin{table}[ht]
\begin{center}
\resizebox{\columnwidth}{!}{%
\begin{tabular}{cccc}
\hline
\textbf{Method}
& \textbf{+train ex.}
& \textbf{\prauc (dev)} 
& \textbf{\prauc (test)} 
\\
\hline
\hline
\albert (\squadtwo)                & 0     & 80.25 & 77.78 \\
+\bioasq                             & 2,848 & 89.57 & 76.78 \\
\hline
+ \wtv      & 12,848 & \textbf{95.60} & 84.99  \\
+ \context  & 19,276 & 93.98 & 83.54  \\
+ \biomrc   & 29,276 & 94.27 & 85.18 \\
+ \btr      & 44,869 & 93.44 & 83.97  \\
\hline

\end{tabular}
}
\vspace*{-3mm}
\caption{Results using a combination of Context Increasing and \wtv data augmentation.}
\vspace*{-5mm}
\label{tab:ensemble_res}
\end{center}
\end{table}

\section{Examples of Artificial Data}
\label{sec:examplesAppendix}

\subsection{\biomrc}
Table~\ref{tab:examples_biomrc} presents training instances generated from the \biomrc dataset. Each instance is a triple containing a cloze-style question, a snippet, and the span of the snippet answering a question. This is very similar to the \squad setting which we have adopted in our experiments (see Section~\ref{sec:data}).

\subsection{Back-translation}
Tables~\ref{tab:examples_BT_question} and~\ref{tab:examples_BT_snippet} show training instances generated via back-translation of \bioasq questions and snippets, respectively. The back-translated questions and snippets retain the semantics of the original ones while adding diversity to the training set.

\subsection{Information Retrieval}
Table~\ref{tab:examples_ir} contains training instances generated via Information Retrieval. A \bioasq question is used as a query in a search engine to retrieve \pubmed documents (abstracts and titles). From the retrieved documents all the snippets containing the answer are extracted and used to generate new training triples. Note that although a retrieved snippet may contain the entity that answers the \bioasq question, it is not always evident that it answers the question, e.g., it may answer another question as is the case in the instance with id 29767248. 

\subsection{Word Substitution}
Tables~\ref{tab:examples_w2v} and~\ref{tab:examples_biolm} presents examples generated via word substitution based on \wtv and \biolm respectively. Although some substitutions may induce noise, the generated snippets tend to retain the semantics of the original ones and add diversity to the training set.

\subsection{Question Generation}
Tables~\ref{tab:examples_qg_bioasq} and~\ref{tab:examples_qg_pubmed} show examples generated via Question Generation using \bioasq snippets and random snippets from random \pubmed articles respectively. Although, the generated triples introduce diverse answers they contain rather simplistic questions which are not indicative of the specialized questions found in \bioasq.

\subsection{Additional Context}
Table~\ref{tab:examples_add_context} contains examples generated by adding context to the original \bioasq snippets. The additional context provides additional information that helps the model to better distinguish relevant and irrelevant parts of the original snippet.

\section{Computing Infrastructure}
All of our experiments run on a titan-X GPU with 12GB of Memory while all code was compiled for CUDA Version 10.2.
The personal computer we used offers 32GB of DDR4-RAM Memory and a 6-core Intel(R) Core(TM) i7-5820K CPU.

\section{Hyper-parameter tuning}

The random seed in all experiments was set to $1$. For data augmentation through Information Retrieval (\ir), we use an ElasticSearch cluster to retrieve relevant abstracts using BM25 with default parameters.

Due to computational and time restrictions, hyper-parameter tuning was performed with grid-search by training on the original 2,848 \bioasq examples (Table \ref{tab:baselines}), i.e., without data augmentation, and evaluating on the development data. The `best' hyper-parameter values were then used in all the augmentation experiments. The hyper-parameter search space (48 settings) and the selected values can be seen in Table~\ref{tab:hyperparameters}.

\begin{table}[ht]
\begin{center}
\resizebox{\columnwidth}{!}{%
\begin{tabular}{ccc}
\hline
\textbf{Hyperparameter}
& \textbf{choices}
& \textbf{best dev. setting} 
\\
\hline
\hline
Random Seed      & \{1\}          & 1   \\
\mlp Hidden Size & \{50, 100\}    & 100 \\
Total Epochs     & \{50, 100\}    & 50  \\
Patience         & \{5\}          & 5   \\
Monitor Score    & \{\auc, loss\} & \auc\\
Learning Rate    & \{0.1, 0.01, 2e-5, 5e-5 \} & 5e-5\\
Weight Decay     & \{0.01\}       & 0.01   \\ 
Warmup Steps     & \{0\}          & 0   \\ 
Batch Size       & \{16, 8\}      & 16  \\ 
\hline
\end{tabular}
}
\caption{Hyper-parameter search space and selected values. We performed a grid-search on a total of $48$ different settings. The best choices per hyper-parameter can be seen in the last column.}
\label{tab:hyperparameters}
\end{center}
\end{table}

\begin{table*}[ht]
    \centering
    \resizebox{\textwidth}{!}
    {
    \begin{tabular}{|p{0.15\textwidth} | p{0.85\textwidth }|}
    \hline
    \multicolumn{2}{|c|}{\textbf{\da with instances from \biomrc}}\\ 
    \hline
    ID  & Instance \\ 
    \hline
    16061304
    & \textbf{\biomrc question:} Prognosis of 6644 resected [MASK] in Japan: a Japanese lung cancer registry study.\\
    &\textbf{\biomrc snippet:} Otherwise, the present TNM staging system seemed to well characterize the stage-specific prognosis in non-small cell lung cancer .\\
    &\textbf{\biomrc answer:} non-small cell lung cancer \\
    \hline
    19823942
    & \textbf{\biomrc question:} Systolic versus diastolic cardiac function variables during [MASK] treatment for breast cancer .\\
    &\textbf{\biomrc snippet:} epirubicin induces considerable decrease in left ventricular ejection fraction and a high risk of CHF. \\
    &\textbf{\biomrc answer:} epirubicin \\
    \hline
    22457372
    & \textbf{\biomrc question:} Pre-operative education and counselling are associated with [MASK] following carotid endarterectomy: a randomized and open-label study.\\
    &\textbf{\biomrc snippet:} AIM: To investigate the effect of pre-operative visits and counselling by intensive care unit ( intensive care unit ) nurses on Patients 's anxiety symptoms following carotid endarterectomy. \\
    &\textbf{\biomrc answer:} anxiety symptoms \\
    \hline
    \end{tabular}
    }
    \caption{Training instances extracted from \biomrc. Each instance is a triple containing a cloze-style question, a snippet, and the span of the snippet answering the question.}
    \label{tab:examples_biomrc}
\end{table*}

\begin{table*}[ht]
    \centering
    \resizebox{\textwidth}{!}
    {
    \begin{tabular}{|p{0.15\textwidth} | p{0.85\textwidth }|}
    \hline
    \multicolumn{2}{|c|}{\textbf{\da via question back-translation}} \\ 
    \hline
    ID  & Instance \\ 
    \hline
    8699317
    &\textbf{Pivot language:} French\\
    &\textbf{\bioasq question:} Which is the gene mutated in type 1 neurofibromatosis?\\
    &\textbf{Back-translated Question:} What is the mutated gene in type 1 neurofibromatosis?\\
    &\textbf{\bioasq snippet:} An NF1 gene was identified as a gene whose loss of function causes an onset of human disorder, neurofibromatosis type I.\\
    &\textbf{\bioasq answer:} NF1 \\
    \hline
    11816795
    &\textbf{Pivot language:} Spansih\\
    &\textbf{\bioasq question:} Which is the primary protein component of Lewy bodies?\\
    &\textbf{Back-translated question:} What is the main protein component of Lewy bodies?\\
    &\textbf{\bioasq snippet:} The protein alpha-synuclein appears to be an important structural component of Lewy bodies, an observation spurred by the discovery of point mutations in the alpha-synuclein gene linked to rare cases of autosomal dominant PD.\\
    &\textbf{\bioasq answer:} alpha-synuclein\\
    \hline
    3056562
    &\textbf{Pivot language:} German\\
    &\textbf{\bioasq question:} Which type of urinary incontinence is diagnosed with the Q tip test?\\
    &\textbf{Back-translated question:} What type of urinary incontinence does the Q tip test diagnose?\\
    &\textbf{\bioasq snippet:} Simple clinical tests for support of the urethrovesical junction, such as the Q tip test, are non-specific in patients with stress urinary incontinence.\\
    &\textbf{\bioasq answer:} stress urinary incontinence\\
    \hline
    \end{tabular}
    }
    \caption{Training instances generated via back-translation of \bioasq questions using French, Spanish, and German as a pivot language. A generated instance contains a back-translated question and the corresponding \bioasq snippet and answer.}
    \label{tab:examples_BT_question}
\end{table*}

\begin{table*}[ht]
    \centering
    \resizebox{\textwidth}{!}
    {
    \begin{tabular}{|p{0.15\textwidth} | p{0.85\textwidth }|}
    \hline
    \multicolumn{2}{|c|}{\textbf{\da via snippet back-translation}} \\ 
    \hline
    ID  & Instance \\ 
    \hline
    8699317
    &\textbf{Pivot language:} French\\
    &\textbf{\bioasq question:} Which is the protein implicated in Spinocerebellar ataxia type 3?\\
    &\textbf{\bioasq snippet:} Ataxin-3 (AT3) is the protein that triggers the inherited neurodegenerative disorder spinocerebellar ataxia type 3 when its polyglutamine (polyQ) stretch close to the C-terminus exceeds a critical length \\
    &\textbf{Back-translated snippet:} Ataxin-3 (AT3) is the protein that triggers spinocerebellar ataxia type 3 in inherited neurodegenerative disorder when its polyglutamine (polyQ) stretches near the C-terminus exceeds a critical length.\\
    &\textbf{\bioasq answer:} Ataxin-3\\
    \hline
    16232326
    &\textbf{Pivot language:} Spanish\\
    &\textbf{\bioasq question:} Which gene is responsible for the development of Sotos syndrome?\\
    &\textbf{\bioasq snippet:} Haploinsufficiency of the NSD1 gene has been implicated as the major cause of Sotos syndrome, with a predominance of microdeletions reported in Japanese patients\\
    &\textbf{Back-translated snippet:} NSD1 gene haploinsufficiency has been implicated as the main cause of Sotos syndrome, with a predominance of microdeletions reported in Japanese patients.\\
    &\textbf{\bioasq answer:} NSD1 gene\\
    \hline
    11154546
    &\textbf{Pivot language:} German\\
    &\textbf{\bioasq question:} Abnormality in which vertebral region is important in the Bertolotti's syndrome?\\
    &\textbf{\bioasq snippet:} Repeated fluoroscopically guided injections implicated a symptomatic L6-S1 facet joint contralateral to an anomalous lumbosacral articulation.\\
    &\textbf{Back-translated snippet:} Repeated fluoroscopic injections implied a symptomatic L6-S1 facet joint contralateral to an abnormal lumbosacral articulation.\\
    &\textbf{\bioasq answer:} lumbosacral\\
    \hline
    \end{tabular}
    }
    \caption{Training instances generated via back-translation of \bioasq snippets using French, Spanish, and German as a pivot language. A generated instance contains a back-translated snippet and the corresponding \bioasq question and answer.}
    \label{tab:examples_BT_snippet}
\end{table*}

\begin{table*}[ht]
    \resizebox{\textwidth}{!}
    {
    \centering
    \begin{tabular}{|p{0.15\textwidth} | p{0.85\textwidth}|}
    \hline
    \multicolumn{2}{|c|}{\textbf{\da via Information Retrieval}} \\ 
    \hline
    ID  & Instance \\ 
    \hline
    25941473
    & \textbf{\bioasq question:} Which is the neurodevelopmental disorder associated to mutations in the X- linked gene mecp2?\\
    &\textbf{Retrieved snippet:} Genotype-specific effects of Mecp2 loss-of-function on morphology of Layer V pyramidal neurons in heterozygous female Rett syndrome model mice. \\
    &\textbf{\bioasq answer:} rett syndrome \\
    \hline
    28708333
    & \textbf{\bioasq question:} Which is the molecular target of the immunosuppressant drug Rapamycin? \\
    &\textbf{Retrieved snippet} Conversion from calcineurin inhibitors to mTOR inhibitors as primary immunosuppressive drugs in pediatric heart transplantation.\\
    &\textbf{\bioasq answer:} mtor \\
    \hline
    29767248
    & \textbf{\bioasq question:} What is the target of the drug Olaparib?\\
    &\textbf{Retrieved snippet:} Mechanistically, dual blockade of PI3K and PARP in ARID1A-depleted gastric cancer cells significantly increased apoptosis detected by flow cytometry, and induced DNA damage by immunofluorescent staining.\\
    &\textbf{\bioasq answer:} parp \\
    \hline
    \end{tabular}
    }
    \caption{Training instances generated via \ir. A \bioasq question is used as the query to retrieve \pubmed documents. For each snippet of the retrieved documents that contains the answer, we generate a new training triplet consisting of the \bioasq question, the snippet and the \bioasq answer.}
    \label{tab:examples_ir}
\end{table*}

\begin{table*}[ht]
    \centering
    \resizebox{\textwidth}{!}
    {
    \begin{tabular}{|p{0.15\textwidth} | p{0.85\textwidth }|}
    \hline
    \multicolumn{2}{|c|}{\textbf{\da with word substitution based on \wtv}}  \\ 
    \hline
    ID  & Instance \\ 
    \hline
    27965160
    &\textbf{\bioasq question:} Sclerostin regulates what process?\\
    &\textbf{\bioasq snippet:} Sclerostin is a soluble \colorbox{Goldenrod}{antagonist} of Wnt/b-catenin signaling secreted \colorbox{GreenYellow}{primarily} by osteocytes. Current evidence \colorbox{Apricot}{indicates} that sclerostin likely functions as a local/paracrine regulator of bone metabolism rather than as an endocrine hormone.\\
    &\textbf{Snippet after \wtv substitution:} sclerostin is a soluble \colorbox{Goldenrod}{agonist} of wnt-b catenin signaling secreted \colorbox{GreenYellow}{mainly} by osteocytes current evidence \colorbox{Apricot}{suggests} that sclerostin likely functions as a localparacrine regulator of bone metabolism rather than as an endocrine hormone \\
    &\textbf{\bioasq answer:} bone metabolism \\
    \hline
    22003227
    &\textbf{\bioasq question:} Which microRNA is the mediator of the obesity phenotype of patients carrying 1p21.3 microdeletions?\\
    &\textbf{\bioasq snippet:} The study also demonstrated significant enrichment of miR-137 at the synapses of cortical and hippocampal neurons, \colorbox{Goldenrod}{suggesting} a \colorbox{GreenYellow}{role} of miR-137 in regulating local synaptic protein synthesis machinery. CONCLUSIONS: This study showed that dosage effects of MIR137 are associated with \colorbox{Apricot}{1p21.3} microdeletions and \colorbox{Lavender}{may} \colorbox{SeaGreen}{therefore} contribute to the ID phenotype in patients with \colorbox{Tan}{deletions} harbouring this \colorbox{SkyBlue}{miRNA}.\\
    &\textbf{Snippet after \wtv substitution:} the study also demonstrated significant enrichment of mir 137 at the synapses of cortical and hippocampal neurons \colorbox{Goldenrod}{indicating} a \colorbox{GreenYellow}{implication} of mir 137 in regulating local synaptic protein synthesis machinerybrbconclusionsb this study showed that dosage effects of mir137 are associated with \colorbox{Apricot}{2q223} microdeletions and \colorbox{Lavender}{might} \colorbox{SeaGreen}{hence} contribute to the id phenotype in patients with \colorbox{Tan}{microinsertions} harbouring this \colorbox{SkyBlue}{micro-rna}\\
    &\textbf{\bioasq answer:} MIR137 \\
    \hline
    21546092 &\textbf{\bioasq snippet:} Beck's Medical Lethality Scale (BMLS) was administered to \colorbox{Goldenrod}{assess} the degree of medical injury, and the SAD \colorbox{GreenYellow}{PERSONS} \colorbox{Apricot}{mnemonic} scale was \colorbox{Lavender}{used} to evaluate suicide risk.\\
    &\textbf{\bioasq question:} What is evaluated with the SAD PERSONS scale?\\
    &\textbf{Snippet after \wtv substitution:} becks medical lethality scale bmls was administered to \colorbox{Goldenrod}{evaluate} the degree of medical injury and the sad \colorbox{GreenYellow}{people} \colorbox{Apricot}{domain-general} scale was \colorbox{Lavender}{utilized} to investigate suicide risk\\
    &\textbf{\bioasq answer:} suicide risk \\
    \hline
    \end{tabular}
    }
    \caption{Training instances generated via word substitution based on \wtv. We randomly select at most 10 words of a \bioasq snippet and substitute each word $w_i$ with its most similar word $w_j$ from the vocabulary of the \wtv model. Highlights of the same color indicate substituted words and the corresponding substitutions.}
    \label{tab:examples_w2v}
\end{table*}

\begin{table*}[ht]
    \centering
    \resizebox{\textwidth}{!}
    {
    \begin{tabular}{|p{0.15\textwidth} | p{0.85\textwidth }|}
    \hline
    \multicolumn{2}{|c|}{\textbf{\da with word substitution based on \biolm}}  \\ 
    \hline
    ID  & Instance \\ 
    \hline
    22140526
    &\textbf{\bioasq question:} Which gene is responsible for red hair?\\
    &\textbf{\bioasq snippet:} The association signals at the MC1R \colorbox{Goldenrod}{gene} \colorbox{GreenYellow}{locus} from CDH were \colorbox{Apricot}{uniformly} more significant than traditional \colorbox{Lavender}{GWA} analyses. The \colorbox{SeaGreen}{CDH} \colorbox{Tan}{test} will contribute towards \colorbox{SkyBlue}{finding} \colorbox{Thistle}{rare} \colorbox{RedOrange}{LOF} variants in GWAS and sequencing studies.\\
    &\textbf{\bioasq snippet after \biolm substitution:} The association signals at the MC1R \colorbox{Goldenrod}{1} \colorbox{GreenYellow}{identified} from CDH were \colorbox{Apricot}{significantly} more significant than traditional \colorbox{Lavender}{association} analyses. The \colorbox{SeaGreen}{proposed} \colorbox{Tan}{findings} will contribute towards \colorbox{SkyBlue}{detecting} \colorbox{Thistle}{novel} \colorbox{RedOrange}{risk} variants in GWAS and sequencing studies.\\
    &\textbf{\bioasq answer:} MC1R \\
    \hline
    26917818
    &\textbf{\bioasq question:} Dinutuximab is used for treatment of which disease?\\
    &\textbf{\bioasq snippet:} CONCLUSIONS Dinutuximab is the first \colorbox{Goldenrod}{anti-GD2} monoclonal antibody approved in combination with \colorbox{GreenYellow}{GM-CSF,} IL-2, and \colorbox{Apricot}{RA} for maintenance treatment of pediatric patients with high-risk neuroblastoma who achieve at least a partial response to \colorbox{Lavender}{first-line} multiagent, \colorbox{SeaGreen}{multimodality} therapy.\\
    &\textbf{\bioasq snippet after \biolm substitution:} CONCLUSIONS Dinutuximab is the first \colorbox{Goldenrod}{human} monoclonal antibody approved in combination with \colorbox{GreenYellow}{recombinant} IL-2, and \colorbox{Apricot}{dexamethasone} for maintenance treatment of pediatric patients with high-risk neuroblastoma who achieve at least a partial response to \colorbox{Lavender}{prior} multiagent, \colorbox{SeaGreen}{standard} therapy.\\
    &\textbf{\bioasq answer:} neuroblastoma \\
    \hline
    27789693
    &\textbf{\bioasq question:} Which database associates human noncoding SNPs with their three-dimensional interacting genes?\\
    &\textbf{\bioasq snippet:} 3DSNP: a \colorbox{Goldenrod}{database} for linking \colorbox{GreenYellow}{human} \colorbox{Apricot}{noncoding} SNPs to their three-dimensional \colorbox{Lavender}{interacting} \colorbox{SeaGreen}{genes}.\\
    &\textbf{\bioasq snippet after \biolm substitution:} 3DSNP: a \colorbox{Goldenrod}{method} for linking \colorbox{GreenYellow}{functional} \colorbox{Apricot}{GWAS} SNPs to their three-dimensional \colorbox{Lavender}{structural} \colorbox{SeaGreen}{structures}\\
    &\textbf{\bioasq answer:} 3DSNP \\
    \hline
    \end{tabular}
    }
    \caption{Training instances generated via word substitution based on \biolm.We randomly select at most 10 words of a \bioasq snippet and we substitute each word $w_i$ with the most probable word $w_j$ suggested by \biolm after masking $w_i$. Highlights of the same color indicate substituted words and the corresponding substitutions.}
    \label{tab:examples_biolm}
\end{table*}

\begin{table*}[ht]
    \centering
    \resizebox{\textwidth}{!}
    {
    \begin{tabular}{|p{0.15\linewidth} | p{0.85\linewidth }|}
    \hline
    \multicolumn{2}{|c|}{\textbf{\da via Question Generation using \bioasq snippets}} \\ 
    \hline
    ID  & Instance \\ 
    \hline
    21159650
    &\textbf{Generated question:} What enzyme inhibits cullin-RING E3 ubiquitin ligases?\\
    &\textbf{\bioasq snippet:} MLN4924 is a first-in-class experimental cancer drug that inhibits the NEDD8-activating enzyme, thereby inhibiting cullin-RING E3 ubiquitin ligases and stabilizing many cullin substrates \\
    &\textbf{Generated answer:} NEDD8\\
    \hline
    17333537
    &\textbf{Generated question:} What type of RNA triggers silencing of inactivation in eutherian mammals?\\
    &\textbf{\bioasq snippet:} In eutherian mammals X inactivation is regulated by the X-inactive specific transcript (Xist), a cis-acting non-coding RNA that triggers silencing of the chromosome from which it is transcribed\\
    &\textbf{Generated answer:} chromosome\\
    \hline
    16800744
    &\textbf{Generated question:} What is the human tissue kallikrein family of?\\
    &\textbf{\bioasq snippet:} The human tissue kallikrein family of serine proteases (hK1-hK15 encoded by the genes KLK1-KLK15) is involved in several cancer-related processes.\\
    &\textbf{Generated answer:} serine proteases\\
    \hline
    \end{tabular}
    }
    \caption{Training instances generated using \tfive. Given a \bioasq snippet \tfive selects a span of the snippet and generates a question that can be answered by that span. We select spans different than the ones used in \bioasq.}
    \label{tab:examples_qg_bioasq}
\end{table*}

\begin{table*}[ht]
    \centering
    \resizebox{\textwidth}{!}
    {
    \begin{tabular}{|p{0.15\linewidth} | p{0.85\linewidth }|}
    \hline
    \multicolumn{2}{|c|}{\textbf{\da via Question Generation using random snippets from random \pubmed abstracts}} \\ 
    \hline
    ID  & Instance \\ 
    \hline
    26935709
    &\textbf{Generated question:} What can be isolated or in combination with accompanying deformities occurring in the forefoot and/or hindfoot?\\
    &\textbf{\pubmed snippet:} Symptoms can be isolated or in combination with accompanying deformities occurring in the forefoot and/or hindfoot.\\
    &\textbf{Generated answer:} Symptoms \\
    \hline
    29260288 
    & \textbf{Generated question:} What supplementation has been integrated into our practice?\\
    &\textbf{\pubmed snippet:} Vitamin D supplementation has been integrated into our current practice.\\
    &\textbf{Generated answer:} Vitamin D\\
    \hline
    30706485
    & \textbf{Generated question:} What were connected to a volume-cycled ventilator after sedation, analgesia and endotracheal intubation?\\
    &\textbf{\pubmed snippet:} After sedation, analgesia and endotracheal intubation, pigs were connected to a volume-cycled ventilator.\\
    &\textbf{Generated answer:} pigs\\
    \hline
    \end{tabular}
    }
    \caption{Training instances generated using \tfive. Given a random snippet from a random \pubmed article \tfive selects a span of the snippet and generates a question that can be answered by that span.}
    \label{tab:examples_qg_pubmed}
\end{table*}

\begin{table*}[ht]
    \centering
    \resizebox{\textwidth}{!}
    {
    \begin{tabular}{|p{0.15\linewidth} | p{0.85\linewidth }|}
    \hline
    \multicolumn{2}{|c|}{\textbf{\da by adding context}} \\ 
    \hline
    ID  & Instance \\ 
    \hline
    15149039
    &\textbf{\bioasq question:} Which metabolite activates AtxA?\\
    &\textbf{\bioasq snippet:} Transcription of the major Bacillus anthracis virulence genes is triggered by CO2, a signal mimicking the host environment.\\
    &\textbf{\bioasq snippet with additional context:} \hl{Transcription of the major Bacillus anthracis virulence genes is triggered by CO2, a signal mimicking the host environment.} A 182-kb plasmid, pXO1, carries the anthrax toxin genes and the genes responsible for their regulation of transcription, namely atxA and, pagR, the second gene of the pag operon. AtxA has major effects on the physiology of B. anthracis. It coordinates the transcription activation of the toxin genes with that of the capsule biosynthetic enzyme operon, located on the second virulence plasmid, pXO2. In rich medium, B. anthracis synthesises alternatively two S-layer proteins (Sap and EA1).\\
    &\textbf{Answer:} CO2 \\
    \hline
    16757427
    &\textbf{\bioasq question:} What tyrosine kinase, involved in a Philadelphia- chromosome positive chronic myelogenous leukemia, is the target of Imatinib (Gleevec)?\\
    &\textbf{\bioasq snippet:} Imatinib was developed as the first molecularly targeted therapy to specifically inhibit the BCR-ABL kinase in Philadelphia chromosome (Ph)-positive chronic myeloid leukemia (CML).\\
    &\textbf{\bioasq snippet with additional context:} The second generation of BCR-ABL tyrosine kinase inhibitors. \hl{Imatinib was developed as the first molecularly targeted therapy to specifically inhibit the BCR-ABL kinase in Philadelphia chromosome (Ph)-positive chronic myeloid leukemia (CML)}. Because of the excellent hematologic and cytogenetic responses, imatinib has moved toward first-line treatment for newly diagnosed CML. However, the emergence of resistance to imatinib remains a major problem in the treatment of Ph-positive leukemia. Several mechanisms of imatinib resistance have been identified, including BCR-ABL gene amplification that leads to overexpression of the BCR-ABL protein, point mutations in the BCR-ABL kinase domain that interfere with imatinib binding, and point mutations outside of the kinase domain that allosterically inhibit imatinib binding to BCR-ABL.\\
    &\textbf{Answer:} BCR-ABL \\
    \hline
    \end{tabular}
    }
    \caption{Training instances generated by adding context around the original \bioasq snippet. In the generated snippet the original one is highlighted.}
    \label{tab:examples_add_context}
\end{table*}

\end{document}